\documentclass[lettersize,journal]{IEEEtran}

\usepackage{amsmath,amsfonts}
\usepackage{algorithmicx}
\usepackage{algorithm}
\usepackage{array}
\usepackage[caption=false,font=normalsize,labelfont=sf,textfont=sf]{subfig}
\usepackage{textcomp}
\usepackage{stfloats}
\usepackage{url}
\usepackage{verbatim}
\usepackage{graphicx}
\usepackage{cite}
\usepackage[utf8]{inputenc} 
\usepackage[T1]{fontenc}    
\usepackage[colorlinks=true, linkcolor=red,
citecolor=blue,urlcolor=blue]{hyperref}       
\usepackage{url}            
\usepackage{booktabs}       
\usepackage{amsfonts}       
\usepackage{nicefrac}       
\usepackage{microtype}      
\usepackage{bbding}
\usepackage{graphicx}
\usepackage{color}
\usepackage{amsmath,amssymb,amsfonts}
\usepackage{graphicx}
\usepackage{textcomp}
\usepackage{bm}
\usepackage{multirow}
\usepackage{booktabs}
\usepackage{bbding}
\usepackage{mfirstuc}
\usepackage[misc]{ifsym} 
\usepackage{graphics}
\usepackage{makecell}
\usepackage{amsthm}
\usepackage{color}

\usepackage{xspace}
\usepackage{mathrsfs}
\usepackage{algpseudocode}   

\usepackage{accents}
\newlength{\dhatheight}

\newlength\savewidth\newcommand\shline{\noalign{\global\savewidth\arrayrulewidth
  \global\arrayrulewidth 1.25pt}\hline\noalign{\global\arrayrulewidth\savewidth}}

\begin{document}

\title{NBMOD: Find It and Grasp It in Noisy Background}

\author{
Boyuan Cao, Xinyu Zhou, Congmin Guo, Baohua Zhang\textsuperscript{*}, Yuchen Liu, Qianqiu Tan\\
NJAU
}

\maketitle


\section*{\textbf{Abstract}}
    Grasping objects is a fundamental yet important capability of robots, and many tasks such as sorting and picking rely on this skill. The prerequisite for stable grasping is the ability to correctly identify suitable grasping positions. However, finding appropriate grasping points is challenging due to the diverse shapes, varying density distributions, and significant differences between the barycenter of various objects. In the past few years, researchers have proposed many methods to address the above-mentioned issues and achieved very good results on publicly available datasets such as the Cornell dataset and the Jacquard dataset. The problem is that the backgrounds of Cornell and Jacquard datasets are relatively simple - typically just a whiteboard, while in real-world operational environments, the background could be complex and noisy. Moreover, in real-world scenarios, robots usually only need to grasp fixed types of objects. To address the aforementioned issues, we proposed a large-scale grasp detection dataset called \textbf{\textit{NBMOD: Noisy Background Multi-Object Dataset for grasp detection}}, which consists of 31,500 RGB-D images of 20 different types of fruits. Accurate prediction of angles has always been a challenging problem in the detection task of oriented bounding boxes. This paper presents a \textbf{\textit{Rotation Anchor Mechanism}} (\textbf{\textit{RAM}}) to address this issue. Considering the high real-time requirement of robotic systems, we propose a series of lightweight architectures called \textbf{\textit{RA-GraspNet}} (\textbf{\textit{GraspNet with Rotation Anchor}}): \textbf{\textit{RARA}} (\textbf{\textit{network with Rotation Anchor and Region Attention}}), \textbf{\textit{RAST}} (\textbf{\textit{network with Rotation Anchor and Semi Transformer}}), and \textbf{\textit{RAGT}} (\textbf{\textit{network with Rotation Anchor and Global Transformer}}) to tackle this problem. Among them, the RAGT-3/3 model achieves an accuracy of 99\% on the NBMOD dataset. The NBMOD and our code are available at https://github.com/kmittle/Grasp-Detection-NBMOD.

\section{Introduction}
\begin{figure}[htbp]
    \centering
    \includegraphics[width=0.45\textwidth]{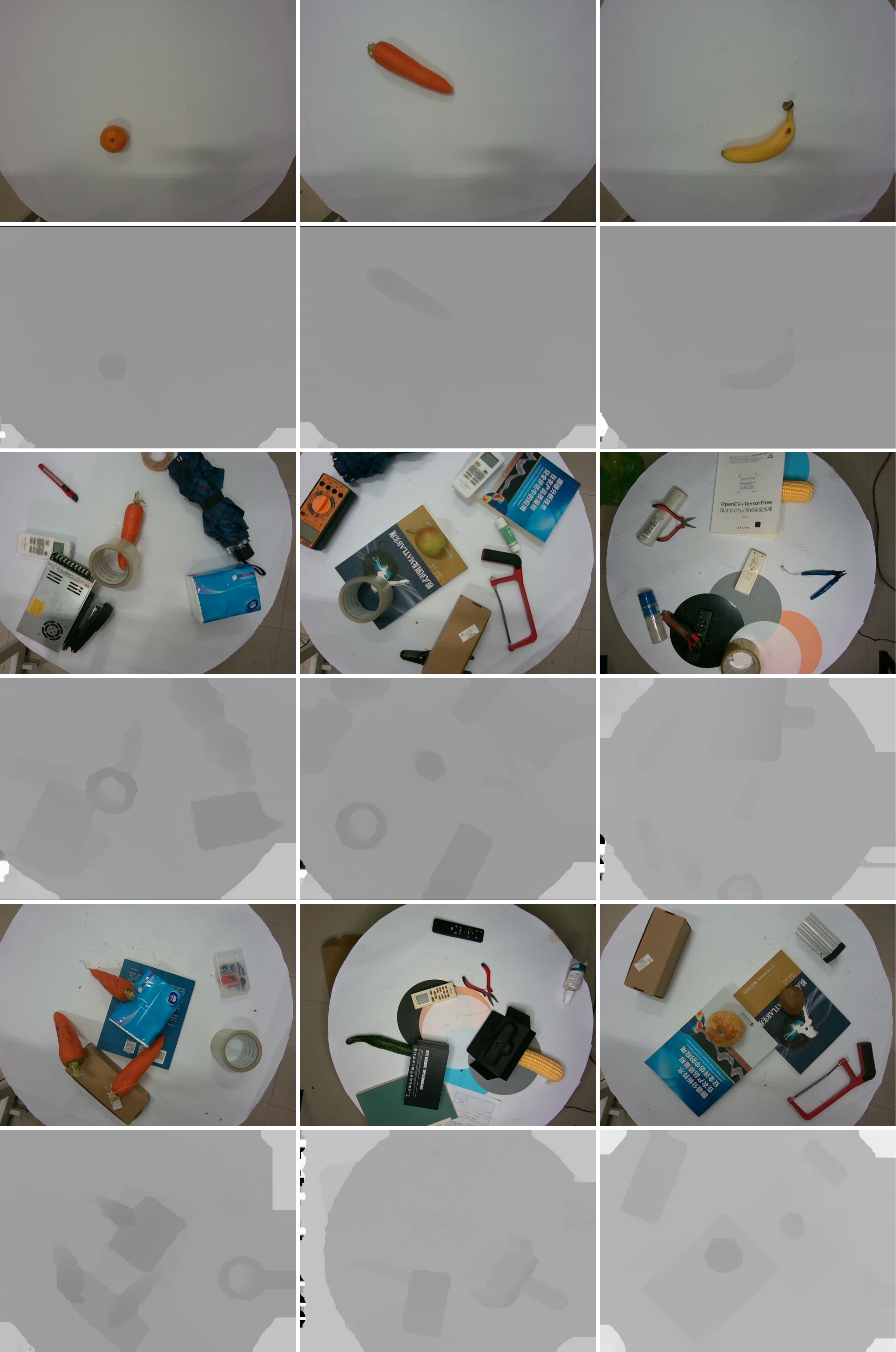}
    \caption{\textbf{The NBMOD dataset consists of RGB and depth images.} The first and second rows depict the RGB and depth images of SSS. The third and fourth rows represent the RGB and depth images of NSS. The fifth and sixth rows display the RGB and depth images of MOS.}
    \label{fig:1}
\end{figure}
In the past decade, the field of machine learning has witnessed revolutionary breakthroughs. In 2012, the deep convolutional neural network AlexNet opened a new chapter in the application of deep learning to large-scale computer vision tasks~\cite{krizhevsky2017imagenet}. Unlike traditional feature engineering methods such as HOG (Histogram of Oriented Gradient), deep neural networks can extract deep semantic information from images, and the features are no longer limited to pixel-level representation~\cite{simonyan2014very, zeiler2014visualizing}. This provides powerful tools for a range of computer vision downstream tasks such as detection and segmentation. After that, a series of object detection algorithms were proposed, such as RCNN, SSD, FPN, RetinaNet, YOLO, etc~\cite{girshick2014rich, girshick2015fast, ren2015faster, redmon2016you, redmon2017yolo9000, redmon2018yolov3, tian2019fcos, liu2016ssd, lin2017feature, lin2017focal, he2015spatial}. These object detection algorithms are crucial for robot grasping detection, as many methods have referenced their ideas. Similar to object detection methods, robot grasp detection methods can also be classified into one-stage and two-stage methods. Two-stage methods in robot grasping detection typically first generate candidate boxes and then select the grasping position with the highest confidence from among the candidate boxes. Single-stage methods typically directly regress the coordinate parameters of the grasp boxes by utilizing the features extracted by the backbone. Therefore, in general, single-stage methods have better real-time performance than two-stage methods. From another perspective, grasp detection algorithms can be divided into anchor-based methods and anchor-free methods. Anchor-free methods have a larger hypothesis space, which allows them to fit more generalized representations when the data size is sufficiently large, but they may perform poorly when the data size is limited. Anchor mechanism endows the model with special inductive bias, which makes it easier for the model to converge with limited data. Anchor-free methods, due to the absence of the inductive bias, can achieve better performance in detecting multi-scale objects. However, robot grasping detection tasks have their unique characteristics, despite their similarity to object detection tasks: 1) target scales usually do not vary significantly; 2) the amount of data is extremely limited; 3) precise angle prediction is challenging. Therefore, anchor-based method is utilized in this paper. Specifically, in the anchor mechanism proposed in this paper - \textbf{the Rotation Anchor Mechanism} (\textbf{RAM}), is mainly designed to address the challenge of accurate angle prediction. Convolutional Neural Networks (CNNs) are a commonly used architecture in computer vision, which possess image-specific inductive biases such as locality, translation invariance, and hierarchy. The problem lies in the fact that in CNNs, the receptive field size is limited in the shallow layers, and the size of the receptive field is closely related to the semantic strength of the representations that a model can learn. The Transformer architecture is good at establishing long-range connections and can capture global information in images even if at shallow layers~\cite{vaswani2017attention, dosovitskiy2020image}. However, due to its lack of image-specific inductive bias, training a Transformer model is often expensive and requires a large amount of data. The architecture of CNN + Transformer can effectively combine the inductive bias of CNN and the global modeling ability of Transformer~\cite{liu2021swin}. Therefore, we chose this architecture as the backbone of our model. Specifically, we choose the Small version of MobileViT as our backbone~\cite{mehta2021mobilevit}.
\par
The NBMOD consists of three parts: 1) the Simple background Single-object Subset (\textbf{SSS}), which contains 13,500 samples, with relatively simple backgrounds; 2) the Noisy background Single-object Subset (\textbf{NSS}), which contains 13,000 samples, with more complex backgrounds such as various daily objects obstructing the detection and occlusion of target objects; and 3) the Multi-Object grasp detection Subset (\textbf{MOS}), which contains 5,000 samples, with each image containing two or more target objects, and the backgrounds are also mostly complex noisy backgrounds. All sample images are RGB-D images, and the labels are in XML format. Some sample images of the NBMOD dataset are shown in Fig.~\ref{fig:1}.
\par
The main contributions of this paper are as follows:
\par
\begin{itemize}
    \item[1)] The proposal of a noisy background multi-object grasping detection dataset called \textbf{NBMOD} containing \textbf{31,500} RGB-D images.
    \item[2)] The \textbf{RA-GraspNet} series detection architectures, including \textbf{RARA}, \textbf{RAST}, and \textbf{RAGT}, were proposed.
    \item[3)] The performance of different detection networks was analyzed, providing references for the study of grasp detection by other researchers.
\end{itemize}
\section{Related Works}
\begin{figure}[htbp]
    \centering
    \includegraphics[width=0.45\textwidth]{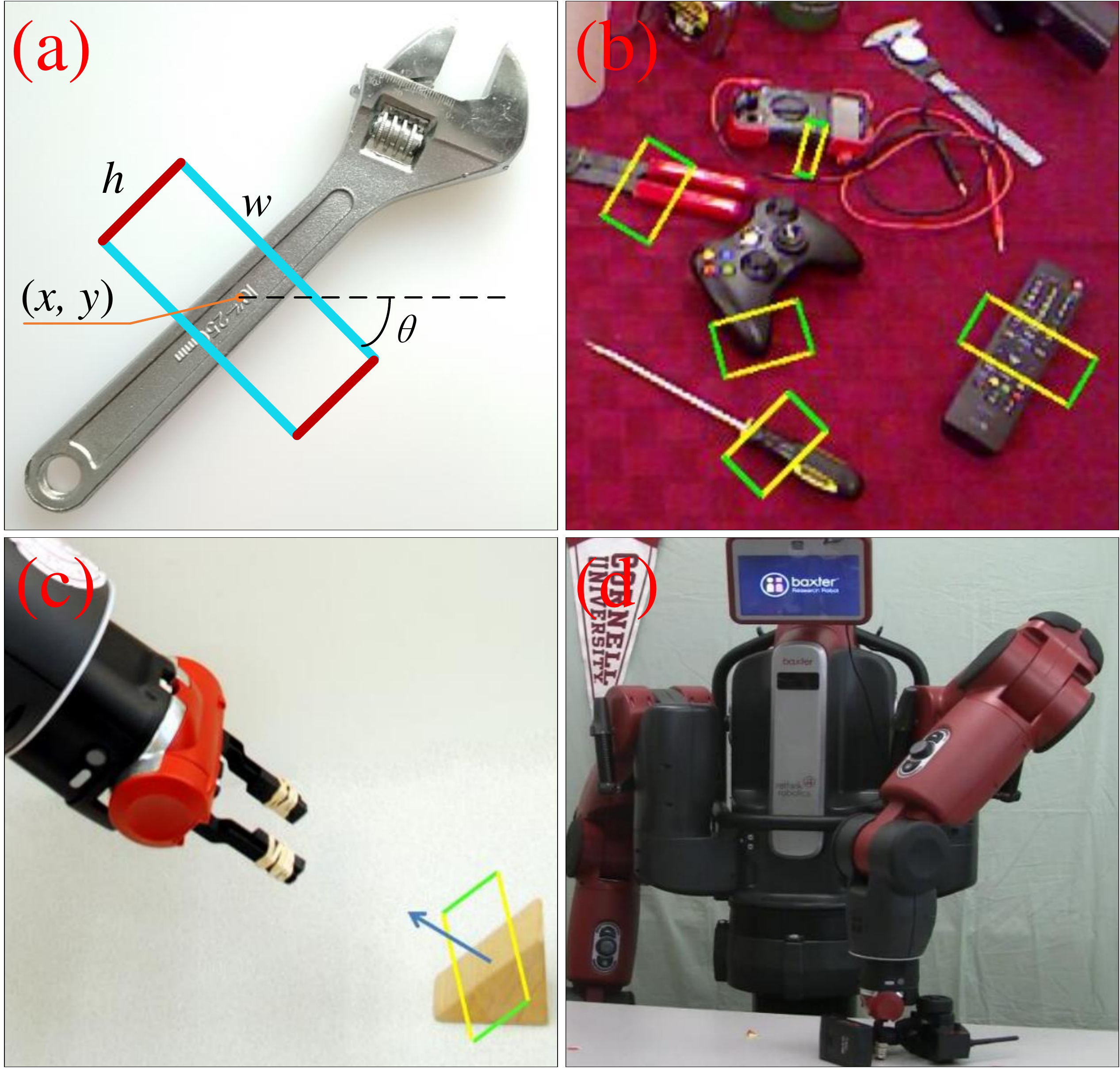}
    \caption{(a) shows an illustration of the five-dimensional grasp representation; (b) displays the detection results achieved by Lenz et al.~\cite{lenz2015deep}; (c) and (d) demonstrate the deployment experiments conducted by Lenz et al.~\cite{lenz2015deep}.}
    \label{fig:2}
\end{figure}

Given an image containing the target object that we want the robot to grasp, we aim for the model to predict the appropriate position for the robot gripper to grasp the object. This position is represented using a five-dimensional grasp representation, as shown in Fig.~\ref{fig:2} (a). In this representation, ($x$, $y$) indicates the center of the oriented bounding box, where $x$ is the horizontal coordinate and $y$ is the vertical coordinate; $w$ represents the width of the rectangle, as indicated by the blue line segment in the figure, which denotes the opening distance of the robot end effector and the direction of closing for grasping; $h$ represents the height of the rectangle, which indicates the width of the robot fingers, as shown by the red line segment in the figure; $\theta$ represents the rotation angle of the rectangle, which is the angle between the w edge and the positive direction of the horizontal axis, and is measured in radians in the label, ranging from 0 to $pi$.

\par
Lenz et al.~\cite{lenz2015deep} were among the earliest scholars who applied deep learning to robot grasping detection. They proposed a detection network consisting of two sub-networks: the first one generated candidate boxes with high confidence, and the second one selected the best box from the samples provided by the first network. Their method achieved an accuracy of 90.3\% on the Cornell dataset when using RGB images. This two-stage network effectively separates the detection task into two parts. The first part is mainly filtering, extracting high-level semantic information from the image, and generating candidate boxes with high confidence. The second part performs fine-grained prediction, similar to refinement. The advantage of this network is high detection accuracy, but the staged prediction increases the complexity of the model and greatly impairs the real-time performance of the model. Similarly, Song et al.~\cite{song2020novel} improved on the Faster R-CNN by using an anchor mechanism to provide the model with prior knowledge for grasp detection. Their method achieved an accuracy of 96.2\% on the Cornell dataset when using RGB images.
\par
 In the realm of single-stage networks, Redmon and Angelova~\cite{redmon2015real} used AlexNet as the backbone and experimented with the following three detection modes: 1) directly regressing grasp coordinates, 2) regressing grasp coordinates while predicting object category, and 3) dividing the image into an $N \times N$ grid and predicting a set of grasp coordinates, confidence, and object category for each grid. Finally, the three methods of direct regression, regression combined with classification, and grid-based prediction achieved accuracy rates of 84.4\%, 85.5\%, and 88\%, respectively, on the Cornell dataset. This indicates that using the additional supervisory signal of classification labels helps the model fit the grasping task, and using grid-based prediction methods can also improve the model's performance. The approach of Joseph Redmon and Anelia Angelova's MultiGrasp network is reasonable, but the main reasons for the low accuracy of their model are twofold: 1) the receptive field of AlexNet is highly limited, and its capability for feature extraction is also insufficient; 2) the model's output is completed using an MLP (Multi-Layer Perceptron), and the process of dimension conversion through the flatten operation may result in the loss of spatial information, which is crucial for grasp detection. Increasing the depth of a neural network is a direct approach to enhancing its nonlinearity, enlarging its receptive field, and strengthening its ability to extract semantic information. However, for a long time, deep networks faced the problem of degradation. In 2015, He et al.~\cite{he2016deep} proposed ResNet, which solved the degradation problem, and since then, ResNet has served as a powerful backbone for many downstream tasks. Kumra and Kanan~\cite{kumra2017robotic} used ResNet-50 as the backbone of their model and conducted two sets of experiments. The first set of experiments involved a single-branch model that takes RGB or RGD images as input. The second set of experiments involved a dual-branch model that takes RGB images and depth images as inputs. In this model, the ResNet networks in the two branches are utilized for extracting features from RGB images and depth images, respectively. Subsequently, the extracted features from both modalities are fused and fed into an MLP to perform regression for estimating grasp coordinates. Among their two approaches, the superior one achieves an accuracy of 89.21\% on the Cornell dataset. Their model uses ResNet-50 as the backbone, which shows improvements over previous work in terms of performance. However, their model still loses a significant amount of spatial information during the global average pooling process, leading to suboptimal results. Similar works have been done by Ribeiro and Grassi, who proposed an end-to-end convolutional neural network architecture that uses CNN to extract features and MLP to predict grasping coordinates~\cite{ribeiro2019fast}. In addition, they applied sliding window screenshot techniques for data augmentation. Their method achieved an accuracy of 94.8\% on the Cornell dataset. Caldera et al.'s work is similar to that of Ribeiro and Grassi, but differs in that they used ResNet-50 as the backbone and achieved a 93.91\% accuracy on the Cornell dataset~\cite{caldera2018robotic}. In order to detect multiple objects, Chu and Vela~\cite{chu2018deep} proposed a multi-object grasp detection model, which decouples angle prediction from directed box center and width-height prediction, and uses MLPs with 20 and 80 neurons respectively to predict grasp oriented box coordinates. This model achieved an accuracy of 96\% on the Cornell dataset. Both Redmon and Angelova, as well as Kumra and Kanan, used anchor-free models in their work. Although they both employed transfer learning with pre-trained weights to improve model performance, the Cornell dataset is simply too small with only 885 images. Data augmentation can address the issue of limited data to some extent, but excessive augmentation may lead to severe feature reuse. Furthermore, the transformations used in data augmentation are assumed to be known and invariant, which hinders the algorithm's ability to handle tasks involving general and unknown transformations that have not been adequately modeled~\cite{dai2017deformable}. Using anchors endows the model with prior knowledge, which can reduce the hypothesis space of the model and decrease its dependence on the amount of data. Zhou et al.~\cite{zhou2018fully} also employed ResNet-50 as the model backbone, but unlike the work of Kumra and Kanan, they used a fully convolutional architecture. This avoided the dimensionality conversion problem caused by using global pooling or flatten operations in the output stage and allowed spatial positional information to be maximally preserved. Their model achieved a 98.87\% accuracy rate on the Cornell dataset. The work of Zhou et al. has achieved great success, but ResNet is considered a rather cumbersome model by today's standards. In ResNet, the signal is primarily transmitted through shortcut branches, and simply increasing model depth by stacking residual blocks to expand the model's receptive field is an inefficient approach~\cite{ding2022scaling}. Similar work was also done by Depierre et al.~\cite{depierre2002optimizing}, who utilized ResNet-50 as the backbone for their network. In addition to the Zhou et al.’s model, they incorporated a scorer mainly composed of MLP into their model. Their model achieved an accuracy rate of 85.74\% on the Jacquard dataset, which outperformed the Zhou et al.’s model that only achieved 81.95\%. Park et al.~\cite{park2018real} used DarkNet-19 with fewer residual blocks as the backbone and proposed a fully convolutional neural network with a skip connection to fuse shallow and deep features for grasp detection. Their model achieved an accuracy of 96.6\% on the Cornell dataset. To achieve pixel-level predictions, some researchers have attempted to use networks with structures similar to UNet for prediction, and have achieved a 98.9\% accuracy on the Cornell dataset~\cite{liu2022pegg}.
\par
An attempt has been made by some scholars to use the Transformer architecture to perform grasp detection tasks. Dong and Yu~\cite{dong2022robotic} employed the Swin Transformer as the backbone and convolutional kernels as the decoder to predict grasp positions. Their model achieved a 98.1\% accuracy rate on the Cornell dataset. In the decoder part, their detector structure has three prediction branches, which is actually unnecessary. In the general object detection task, the multi-branch prediction structure in the form of FPN is used to better address the problem of the model's difficulty in detecting objects of different scales. However, in the robotic grasp detection task, the object size variation is not so drastic, and too many branches actually increase the complexity of post-processing, which adversely affects the real-time performance of the model. Wang et al.~\cite{wang2022transformer} have proposed a Transformer-based architecture, named TF-Grasp, which also adopts window attention to reduce the computation cost of self-attention. In addition, shortcut branches are utilized in TF-Grasp to integrate shallow details with deep semantic information. Their model achieves a precision rate of 97.99\% on the Cornell dataset. While the aforementioned works all employ the five-dimensional grasp representation method, which uses a oriented box to indicate the graspable position and is the most popular approach, there are also some works that have used alternative representation methods, such as line segments, oriented triangles, and oriented circles~\cite{wang2020sgdn, guo2016object, xu2019graspcnn}. Regarding the model input, RGB-D images are the mainstream input signal for grasp detection, but some works have also achieved good results by using signals from point clouds or combining visual and tactile information to complete the grasping task~\cite{mahler2019learning, mahler2017dex, mahler2018dex, guo2017hybrid}.
\par
Compared to previous works, we focus on the recognition and grasping of specific targets in noisy environments. We proposed the \textbf{RA-GraspNet} series of architectures, including \textbf{RARA}, \textbf{RAST} and \textbf{RAGT}, to solve the detection problem. Our experiments show that our models perform well even in environments filled with noise.
\section{Methods}

\subsection{NBMOD}
\begin{figure}[htbp]
    \centering
    \includegraphics[width=0.45\textwidth]{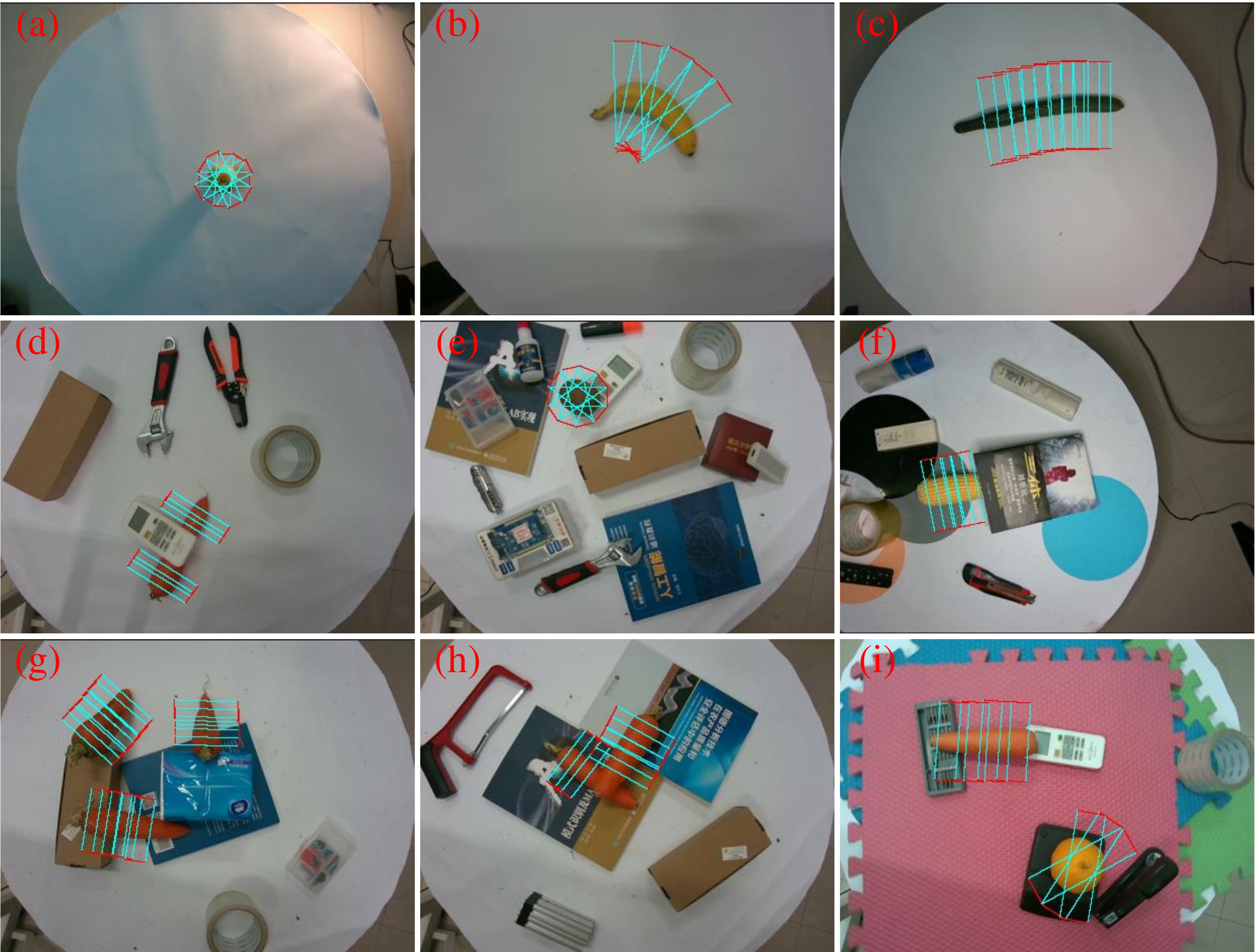}
    \caption{The first row of images shows examples of SSS annotations; the second row shows examples of NSS annotations, and the third row shows examples of MOS annotations.}
    \label{fig:3}
\end{figure}

\subsubsection{\textbf{Acquiring Images}}
The Realsense D455 depth camera was used to capture RGB-D images, and the spatial positions of the target objects in each image are randomized. In the process of collecting noisy scene data, various unrelated office supplies were used to occlude the target object or to introduce interference. During dataset acquisition, the camera was positioned approximately 45-60cm from the working surface. During the process of image acquisition, there exist certain variations in both the background and illumination, as well as the introduction of interferences. During the image acquisition process, approximately 50 samples were used for each type of grasping target.

\subsubsection{\textbf{Image Annotation}}
Rolabelimg was used to annotate suitable grasping positions. There are usually more than one suitable grasping position, and if only one is annotated, it may cause the model to fail to fit a generalized representation for grasping tasks, or even make the model difficult to converge. On the other hand, too few annotations may result in serious misjudgment when evaluating the model, because determining whether a predicted grasping position is correct involves the calculation of IoU (Intersection over Union) of the oriented box (as explained in~\ref{sec:eval}). Therefore, the suitable grasping positions for robots were annotated as much as possible, as shown in Fig.~\ref{fig:3}.
\par
In the presence of obstacles or multiple target objects, some positions are actually not suitable for grasping, and forcing the grasp task may damage the target object. As shown in Fig.~\ref{fig:3} (g), due to the presence of the blue box, the ends of some carrots are actually not suitable for grasping. In Fig.~\ref{fig:3} (h), grasping the overlapping part of the carrots may crush the carrots underneath. In Fig.~\ref{fig:3} (i), the presence of a stapler also obstructed some positions where the oranges could be grasped. We have taken these situations into consideration and have specifically labeled these samples to make the dataset more representative of the actual conditions of robot tasks.

\subsection{Rotation Anchor Mechanism}

\subsubsection{\textbf{Coordinate Regression}}
In our view, the essence of using anchors is to narrow down the hypothesis space of the model. Directly outputting coordinates may lead to a large output range, which can cause significant coordinate deviation in point regression problems such as object detection. Based on anchors, the model regresses not the coordinates themselves, but rather the offset between the expected coordinates and the anchors. Compared to directly regressing coordinate points, regressing offsets can achieve higher precision with the same relative deviation.
\par
In contrast to conventional object detection, predicting the appropriate grasping angle is a challenge in the process of detecting suitable grasping positions. Our proposed  Rotation Anchor Mechanism (RAM) can effectively address this issue. Specifically, we partition the input image into an $N \times M$ grid and the input image is down-sampled $n$ times to generate an $N \times M \times k \times 6$ tensor, where $N$ and $M$ are the height and width of the tensor, $k$ is the number of anchors contained in each grid cell, and each anchor generates a predicted box during the inference stage. The meaning of "6" is the confidence score for each predicted box and the offset of the five coordinate parameters relative to its corresponding anchor. For each anchor, its five coordinate parameters are defined as follows: 1) its $x_{anchor}$ and $y_{anchor}$ values are the horizontal and vertical index values of the grid cell where it is located (the index starts at 0); 2) the $k$ anchors in each grid divide the $0°$ to $180°$ interval into $k$ parts (The model outputs rotation angles in degrees rather than radians for oriented boxes), and the $\theta_{anchor}$ value of the m-th anchor is $(m-1)\times\theta_{margin}$ (m=1, 2, ..., k), where $\theta_{margin}$ is defined as in Eq.~(\ref{eq:1}); 3) we note that the $w$ and the $h$ of different ground truths in the training set vary little, so we take the average of the $w$ and $h$ of all ground truths in the training set as the $w_{anchor}$ and $h_{anchor}$ of the anchor.

\begin{equation}
\theta_{\text {\textit{margin}}}=\frac{180}{k}
\label{eq:1}
\end{equation}

\par
Assuming that the original image size is $H \times W$, after down-sampled by the model, we obtain a tensor of size $N \times M \times k \times 6$, where each grid corresponds to a region of size $x_{margin}\times y_{margin}$ in the original image. The definitions of $x_{margin}$ and $y_{margin}$ are given by Eq.~(\ref{eq:2}).

\begin{equation}
\left\{\begin{array}{l}
x_{\text {\textit{margin} }}=\frac{\textit{H}}{\textit{N}} \\
\\
y_{\text {\textit{margin} }}=\frac{\textit{W}}{\textit{M}}
\end{array}\right.
\label{eq:2}
\end{equation}

\par
The final prediction result $\hat{g}=\{\hat{x}, \hat{y}, \widehat{w}, \hat{h}, \hat{\theta}\}$ can be obtained based on the compensation values predicted by the model and the coordinate parameter values of the anchor, as shown in Eq.~(\ref{eq:3}).

\begin{equation}
\left\{\begin{array}{l}
\hat{x}=\left(x_{\text {\textit{anchor }}}+\sigma\left(t_x\right)\right) \times x_{\text {\textit{margin} }} \\
\hat{y}=\left(y_{\text {\textit{anchor} }}+\sigma\left(t_y\right)\right) \times y_{\text {\textit{margin} }} \\
\hat{w}=w_{\text {\textit{anchor }}} \times e^{t_w} \\
\hat{h}=h_{\text {\textit{anchor }}} \times e^{t_h} \\
\hat{\theta}=\theta_{\text {anchor }}+\sigma\left(t_\theta\right) \times \theta_{\text {\textit{margin} }}
\end{array}\right.
\label{eq:3}
\end{equation}

\par
The values $t_x, t_y, t_w, t_h$, and $t_\theta$, represent the offsets of the model output relative to the anchor. The function $\sigma(x)$ is the sigmoid activation function, which is defined as shown in Eq.~(\ref{eq:4}).

\begin{equation}
\sigma(x)=\frac{1}{1+e^{-x}}
\label{eq:4}
\end{equation}

\par
By normalizing the output offset of the model to between 0 and 1, $\sigma(x)$ greatly reduces the decision space of the model, making its predictions more accurate and training easier to converge.

\subsubsection{\textbf{Matching Strategy}}

In conventional object detection, only the coordinate parameter deviation of positive samples needs to be included in the loss for training, and for negative samples, only their confidence deviation is included in the loss for training. The same loss calculation method is used in grasp detection. The model output tensor dimension is $N \times M \times k \times 6$. Each grid cell produces k predicted boxes, each of which is refined from a specific anchor, but not every predicted box is a positive sample. We stipulate that positive sample predicted boxes must simultaneously satisfy the following two conditions: 1) there is at least one ground truth center in the grid cell corresponding to the anchor of the predicted box; 2) the difference between the ground truth angle and the $\theta_{anchor}$ of the anchor corresponding to the predicted box is less than $\theta_{margin}$. All samples except positive samples are negative samples. The confidence label value for all positive samples is 1, and the confidence label value for all negative samples is 0.
Compared with the previous anchor mechanism~\cite{song2020novel, zhou2018fully}, the proposed rotation anchor mechanism in this paper scales the model's decision space using $\sigma(x)$. This ensures that: 1) the predicted center of the oriented box will not exceed the range of the grid cell it belongs to; 2) the angle deviation will not exceed $\theta_{margin}$. This allows the model to perform more accurate localization. The \textbf{RAM} is, in fact, a \textbf{\textit{plug-and-play}} generic module that can be used for various tasks that involve oriented bounding box localization, such as remote sensing localization. For different tasks, multiple branches can be designed to detect objects of different scales. It is only necessary to design appropriate anchor sizes for the branches that detect objects of different scales.

\subsection{Models}

Conventional object detection algorithms, such as SSD, RetinaNet, and YOLO series algorithms, employ multi-branch detection heads with structures similar to FPN. In neural networks, deeper layers possess larger receptive fields and often yield more semantically rich representations. However, they may also lose image detail information due to multiple filtering and downsampling processes, making them suitable for detecting large-scale objects. In contrast, shallow layer representations, despite having weaker semantic strength, retain more image details due to fewer downsampling operations, making them suitable for detecting small-scale objects. Combining deep and shallow layer features for multi-scale object detection can better ensure the model's detection performance. However, in the context of robotic grasping detection, the situation is somewhat different. Typically, the working platform distance faced by robots during grasping tasks is fixed, and the size differences of the target objects are not significant. This determines that the captured images from the camera will not exhibit drastically varying scales of the target objects. The use of multi-branch detection heads comes at the expense of model parallelism to some extent. Generally speaking, the more branches a network has, the higher the degree of fragmentation, resulting in slower model inference speed~\cite{ma2018shufflenet}. However, robotic grasping and sorting systems should be real-time systems, where better synchronization and detection efficiency are more desirable. To expedite the process, only a single-branch detection head is employed in all models.

\subsubsection{\textbf{\textit{The Architecture of RAST}}}

\begin{figure*}[htbp]
    \centering
    \includegraphics[width=0.9\textwidth]{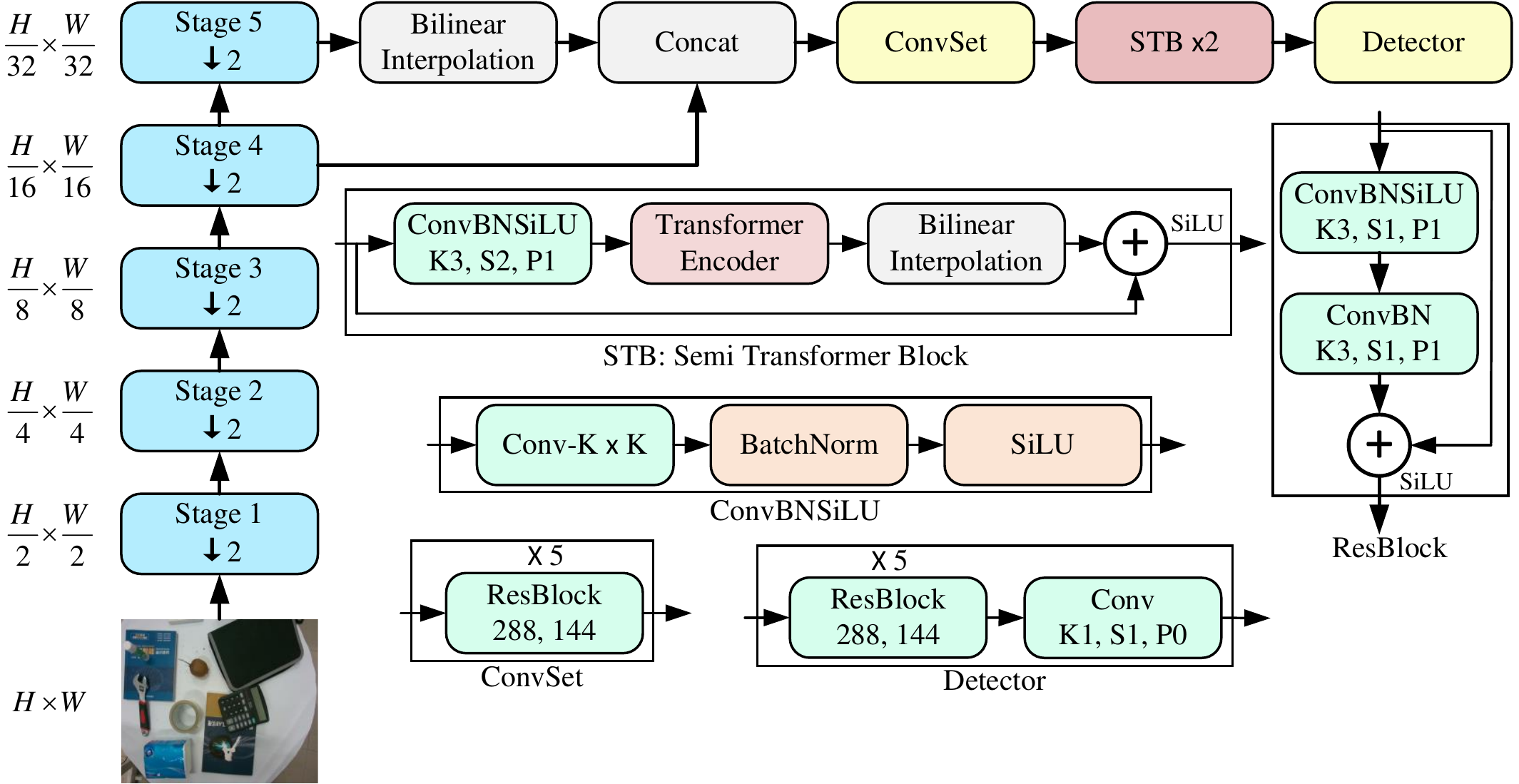}
    \caption{\textbf{The architecture of RAST}. Stages 1 to 5 represent the 5 downsampling stages of the backbone. "↓2" indicates that this module downsamples the feature map by a factor of 2. In the ConvSet structure, 288 represents the number of input channels and 144 represents the number of output channels of the first convolutional layer in the ResBlock. In the ConvBNSiLU block, S denotes the convolutional stride, K denotes the convolutional kernel size, and P denotes padding.}
    \label{fig:4}
\end{figure*}

To enable more precise detection, we consider using feature maps that have undergone 16 times down-sampling as the output and merging them with feature maps that have undergone 32 times down-sampling to enhance semantic features. Given that the MobileViT employs patch-based Transformer modules, the spatial information is discontinuous during self-attention calculation. To address this issue, we envisage using a global Transformer in the detection head to integrate the overall information. However, the serialization of the feature maps that have undergone 16 times down-sampling results in too many tokens, leading to an increase in computational cost. Therefore, we introduce the \textbf{RAST} (\textbf{network with Rotation Anchor and Semi Transformer}) architecture for efficient detection, as shown in Fig.~\ref{fig:4}. Absolute positional encoding is used in the Transformer encoder. In the Transformer Encoder, we employ a 4-layer, 4-head Transformer architecture. The token dimension used was 288, and during the MLP projection, the token dimension was first projected to 432 and then projected back to 288.
\par
In RAST, the residual mechanism is introduced to ensure that the representation fitted when fusing global features does not degrade. After two computations of self-attention modules consisting of 4 layers, the model has acquired a receptive field that encompasses the global features of the image, enabling the model to better perform the grasping detection task.

\subsubsection{\textbf{\textit{The Architecture of RARA}}}

\begin{figure*}[htbp]
    \centering
    \includegraphics[width=0.95\textwidth]{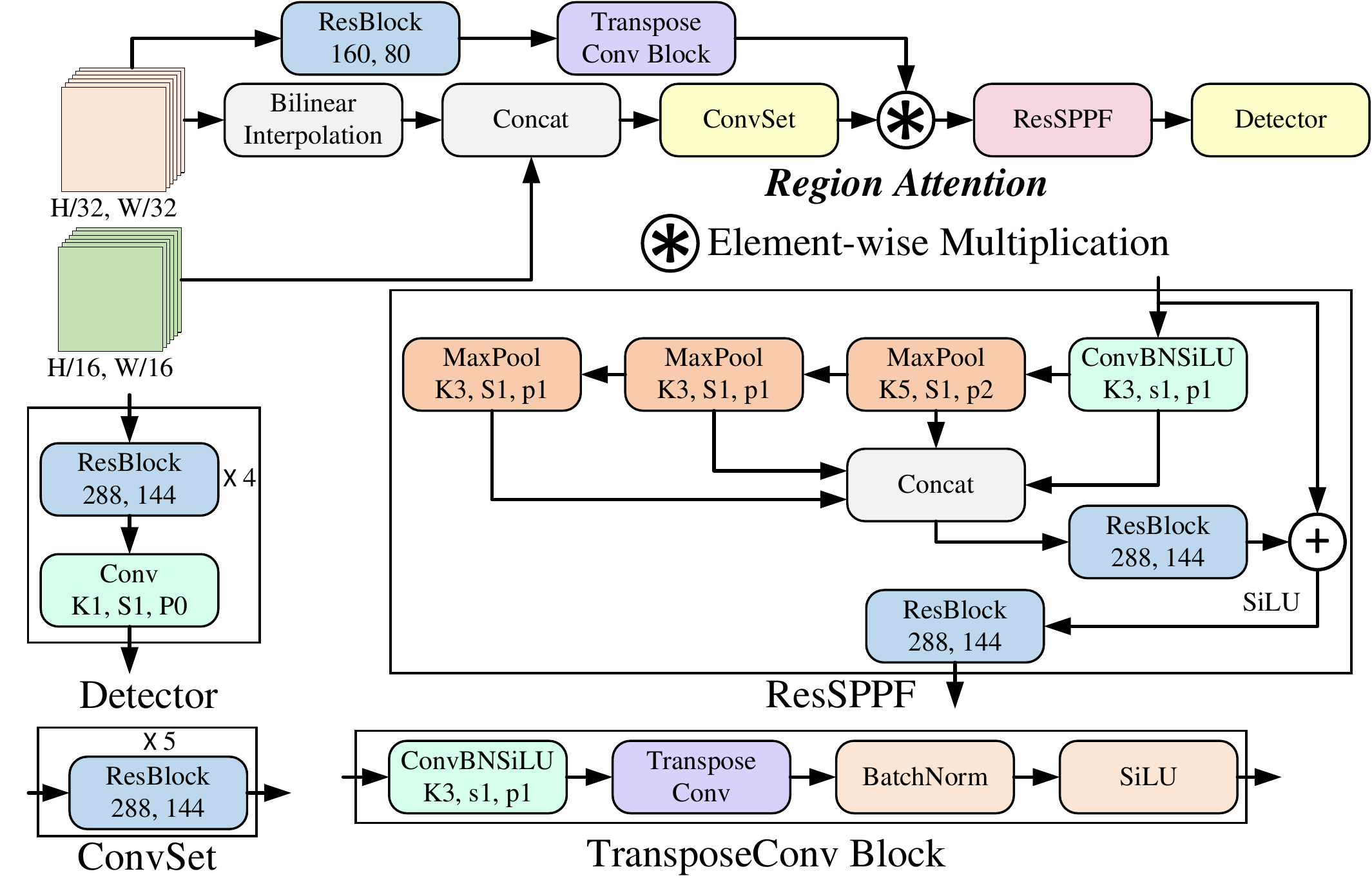}
    \caption{\textbf{The architecture of RARA}. Some structures are the same as those in Fig.~\ref{fig:4}.}
    \label{fig:5}
\end{figure*}

Convolutional computations have undergone multiple optimizations, resulting in faster processing speeds. Therefore, we consider designing an architecture, named \textbf{RARA} (\textbf{network with Rotation Anchor and Region Attention}), that utilizes a pure convolutional decoder. In RARA, we maintain the output size based on feature maps downsampled by a factor of 16. However, the novelty lies in employing a learnable approach to upsample the feature maps downsampled by a factor of 32 within RARA. To accomplish this, we employ transpose convolution during the implementation process. Moreover, considering that feature maps that have undergone 32 times downsampling have stronger semantic features, we propose using the results of transpose convolution to generate dynamic weights to adjust the attention distribution of the model in each grid region of the entire image, which we call region attention. The architecture of RARA is shown in Fig.~\ref{fig:5}.
\par
The convolution layer preceding the transpose convolution is employed to adjust the number of channels to match the output channel of ConvSet, enabling element-wise multiplication, because the transpose convolution solely performs upsampling on the width and height of the feature map. In the conventional SPPF (Spatial Pyramid Pooling - Fast) module, it is necessary to first reduce the channel dimension of the feature map to one-fourth of its original size, and then restore it in subsequent steps. However, this information bottleneck may result in a significant loss of crucial information. Therefore, this paper proposes a \textbf{ResSPPF} (\textbf{Residual Spatial Pyramid Pooling - Fast}) module with residual branches and applies it in RARA to address this issue.

\subsubsection{\textbf{\textit{The Architecture of RAGT}}}

\begin{figure*}[htbp]
    \centering
    \includegraphics[width=0.95\textwidth]{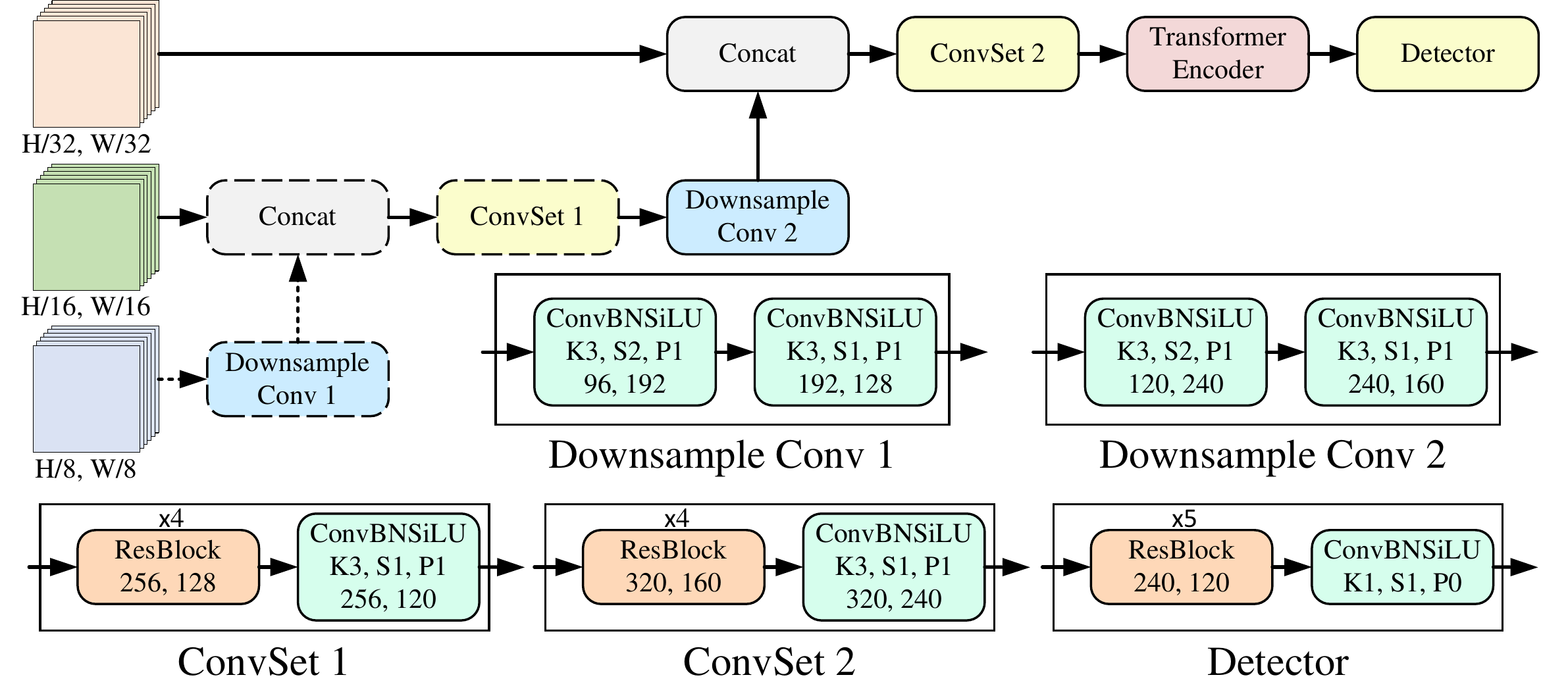}
    \caption{\textbf{The architecture of RAGT.} Some structures are the same as those in Fig.~\ref{fig:4}.}
    \label{fig:6}
\end{figure*}

In the RAST and RARA models, the output is based on feature maps downsampled by a factor of 16. To directly increase the model's receptive field, an alternative approach is to utilize feature maps downsampled by a greater factor for output. In the case of \textbf{RAGT} (\textbf{network with Rotation Anchor and Global Transformer}), the output tensor is derived from feature maps downsampled by a factor of 32. Owing to the increased downsampling rate, the number of tokens for self-attention computation in RAGT is reduced to one-fourth of that in RAST and RARA. Consequently, direct self-attention computation is performed on the global information within the RAGT framework. However, this approach may lead to the loss of fine-grained image details. Therefore, this study proposes a convolution-based FPN structure to restore shallow-level detail information. The architecture of RAGT is illustrated in Fig.~\ref{fig:6}.
\par
The Transformer encoder consists of four layers, with a token dimension of 240 and four attention heads for self-attention. When going through the MLPs, the token dimension is first projected to 480 and then projected back to 240. Two RAGT structures have been devised: 1) The first one is RAGT-3, which incorporates feature maps of three scales. 2) The second one is RAGT-2, which employs feature maps of only two scales. In Fig.~\ref{fig:6}, when the dashed structure is absent, it depicts the RAGT-2 architecture. In RAGT-2, the modules "Downsample Conv 2", "ConvSet 2", and "Detector" exhibit some small differences in terms of channel numbers compared to RAGT-3.

\subsubsection{\textbf{\textit{Loss Function}}}

The loss function is defined by Eq.~(\ref{eq:5}) to (\ref{eq:8}) and comprises three distinct components. The first component corresponds to the loss in confidence scores for positive samples, while the second component pertains to the loss in confidence scores for negative samples. The third component encompasses the loss in coordinate regression for positive samples.

\begin{equation}
\operatorname{\textit{Loss}}_1=-\sum c_p \ln \sigma\left(\hat{c}_p\right)
\label{eq:5}
\end{equation}

\begin{equation}
\operatorname{\textit{Loss}}_2=-\sum\left(1-c_n\right) \ln \left(1-\sigma\left(\hat{c}_n\right)\right)
\label{eq:6}
\end{equation}

\begin{equation}
    \begin{aligned}
    \operatorname{\textit{Loss}}_3=\sum c\left[\left(\hat{t}_x-t_x\right)^2+\left(\hat{t}_y-t_y\right)^2+\right. \\
    \left.\left(\hat{t}_w-t_w\right)^2+\left(\hat{t}_h-t_h\right)^2+\left(\hat{t}_\theta-t_\theta\right)^2\right]
    \end{aligned}
    \label{eq:7}
\end{equation}

\begin{equation}
\text { \textit{Loss} }=\lambda_1 \operatorname{\textit{Loss}}_1+\lambda_2 \operatorname{\textit{Loss}}_2+\lambda_3 \operatorname{\textit{Loss}}_3
\label{eq:8}
\end{equation}

\par
In Eq.~(\ref{eq:5})\textbf{ }to (\ref{eq:8}), the $c$ represents the confidence label value for a given sample. $c_p$ and ${\hat{c}}_p$ correspond to the confidence label value and predicted value for positive samples, respectively. Similarly, $c_n$ and ${\hat{c}}_n$ denote the confidence label value and predicted value for negative samples. The offset parameters, namely $t_x, t_y, t_w, t_h$, and $t_\theta$, represent the label values for the offset values, which can be derived from the ground truth $g=\left\{x,y,w,h,\theta\right\}$ and Eq.~(\ref{eq:3}). The predicted values for the offset parameters are denoted as ${\hat{t}}_x, {\hat{t}}_y, {\hat{t}}_w, {\hat{t}}_h$, and ${\hat{t}}_\theta$. The weight parameters $\lambda_1, \lambda_2$, and $\lambda_3$ are utilized to govern the relative importance of the various components of the loss function.

\subsection{Evaluation Criterion} \label{sec:eval}
For a predicted grasp box $\widehat{G}$, the angular deviation and IoU with the ground truth $G$ are computed. If both inequalities (\ref{eq:9}) and (\ref{eq:10}) are simultaneously satisfied, the predicted grasp box is considered correct~\cite{song2020novel, redmon2015real, kumra2017robotic, zhou2018fully, dong2022robotic, wang2022transformer}.

\begin{equation}
|\theta-\hat{\theta}| \leq 30^{\circ}
\label{eq:9}
\end{equation}

\begin{equation}
\frac{|G \cap \widehat{G}|}{|G \cup \widehat{G}|} \geq 0.25
\label{eq:10}
\end{equation}

\section{Experiments and Analyses}

\subsection{Training Details}

\begin{table}[htbp]
\centering
\caption{Training parameters of each model.}
\begin{tabular}{lccccc}
\shline
Model & Batch size & $\lambda_1$ & $\lambda_2$ & $\lambda_3$ \\
\midrule
RARA-1 & 75 & 2 & 0.003 & 10 \\
RARA-3 & 75 & 2 & 0.003 & 10 \\
RARA-4 & 75 & 2 & 0.003 & 10 \\
RARA-9 & 70 & 2 & 0.002 & 10 \\
RARA-15 & 75 & 2 & 0.0012 & 10 \\
\hline
RAST-1 & 75 & 2 & 0.003 & 10 \\
RAST-2 & 75 & 2 & 0.003 & 10 \\
RAST-3 & 70 & 2 & 0.003 & 10 \\
RAST-9 & 75 & 2 & 0.002 & 10 \\
RAST-15 & 75 & 2 & 0.0014 & 10 \\
\hline
RAGT-2/1 & 80 & 2 & 0.04 & 10 \\
RAGT-2/3 & 80 & 2 & 0.024 & 10 \\
RAGT-2/4 & 80 & 2 & 0.02 & 10 \\
RAGT-2/9 & 75 & 2 & 0.01 & 10 \\
RAGT-2/15 & 70 & 2 & 0.004 & 10 \\
\hline
RAGT-3/1 & 75 & 2 & 0.04 & 10 \\
RAGT-3/3 & 80 & 2 & 0.024 & 10 \\
RAGT-3/4 & 80 & 2 & 0.02 & 10 \\
RAGT-3/9 & 80 & 2 & 0.008 & 10 \\
RAGT-3/15 & 75 & 2 & 0.006 & 10 \\
\shline 
\label{tab:1}
\end{tabular}
\end{table}

We conducted algorithm testing experiments using NBMOD's SSS and NSS. The MOS component was not utilized for testing purposes, as it was primarily designed to support experiments related to robot deployment. The input image size for all models is $416\times416$. The following experiments used a random seed of 42 to split the training set and test set. We partitioned the 13,500 SSS samples into 13,000 training set samples and 500 test set samples. Similarly, the 13,000 NSS samples were divided into 12,500 training set samples and 500 test set samples. Subsequently, we merged the test sets of SSS and NSS. As a result, we obtained a training set comprising 25,500 samples and a test set comprising 1,000 samples. Next, we applied data augmentation techniques to the training set samples using the following methods: 1) clockwise rotation - 90 degrees, 180 degrees, 270 degrees; 2) vertical flipping; 3) horizontal flipping; 4) altering image lighting intensity using PCA (Principal Component Analysis) algorithm; 5) adding Gaussian noise. The experiment employed an A40 GPU with 48GB of storage and a 15-core AMD EPYC 7543 CPU equipped with 80GB of memory. The training optimizer used is AdamW, and no special training tricks were employed in this experiment. The default training parameters of AdamW were utilized ($learning\_rate = 0.001, \beta_1 = 0.9, \beta_2 = 0.99, \varepsilon = {10}^{-8}, weight\_decay = 0.01$). Other training parameters of each model are listed in Table~\ref{tab:1}. For the RARA and RAST models, "RARA-m" indicates that each grid cell in the model has m anchors, while "RAST-n" means that each grid cell in the model has n anchors. For the RAGT model, "RAGT-p/q" means that the model utilizes p different scales of feature maps, and each grid cell in the model contains q anchors.

\subsection{Model Parameters}

Taking RARA-3, RAST-3, RAGT-2/3, and RAGT-3/3 models as representatives, their model parameters are listed in Table~\ref{tab:2}. The difference between models of the same architecture with different numbers of anchors lies only in the output channel number of the final 1x1 convolutional layer. Therefore, the parameter size of models with the same architecture is generally similar. In order to facilitate deployment and accelerate the inference speed of the model, the parameter size of all models does not exceed 20M, and the model's FLOPs are also relatively low.

\begin{table}[htbp]
\centering
\caption{Model parameters.}
\label{tab:quantitative_results_on_25dose_dataset}
\begin{tabular}{lcccc}
\shline
Model & Total parameters & FLOPs & Speed (images per second) \\
\midrule
RARA-3 & 17.96M & 13.45G & 815.2 \\
RAST-3 & 19.64M & 10.55G & 669.6 \\
RAGT-2/3 & 17.94M & \textbf{6.88G} & \textbf{844.0} \\
RAGT-3/3 & \textbf{17.77M} & 8.39G & 742.4 \\
\shline 
\label{tab:2}
\end{tabular}
\end{table}

\subsection{Detection Results and Analysis}

The detection results of each model are presented in Table~\ref{tab:3}. The accuracy of the models did not exhibit an upward trend as the number of anchors increased. We assume that this can be attributed to two factors. The first factor pertains to the necessity of aligning the selection of the number of anchors with the training strategy. When utilizing a higher number of anchors, a more sophisticated training strategy is necessary. However, in this experiment, the models were trained directly with default parameters, without incorporating learning rate decay. By employing a more meticulous training strategy while utilizing a greater number of anchors, it is plausible to achieve improved outcomes. The second reason relates to the exacerbation of the imbalance between positive and negative samples when employing an increased number of anchors. The number of annotated bounding boxes for each image is inherently limited, setting an upper limit on the number of positive samples. Consequently, the introduction of a larger number of anchors generates a substantial amount of unmatched negative samples, thus disrupting the balance and leading to training instability. This further emphasizes the need for a more intricate training strategy when employing a higher number of anchors.

\begin{table}[htbp]
\centering
\caption{The detection results of each model.}
\label{tab:quantitative_results_on_25dose_dataset}
\begin{tabular}{lccccc}
\shline
Architecture & Model & SSS (\%) & NSS (\%) & Mean (\%) \\
\midrule
\multirow{5}{*}{ RARA } & 1 & 94.2 & 87.8 & 91.0 \\
~ & 3 & \textbf{98.2} & \textbf{94.8} & \textbf{96.5} \\
~ & 4 & 90.2 & 60.2 & 75.2 \\
~ & 9 & 95.0 & 83.6 & 89.3 \\
~ & 15 & 90.4 & 88.8 & 89.6 \\
\hline
\multirow{5}{*}{ RAST } & 1 & 95.6 & 60.4 & 78.0 \\
~ & 2 & 95.8 & 92.0 & 93.9 \\
~ & 3 & 96.8 & 91.6 & 94.2 \\
~ & 9 & \textbf{97.6} & \textbf{97.2} & \textbf{97.4} \\
~ & 15 & 95.2 & 89.2 & 92.2 \\
\hline
\multirow{10}{*}{ RAGT } & $2 / 1$ & 98.0 & 96.4 & 97.2 \\
~ & $2 / 3$ & \textbf{98.6} & 98.0 & \textbf{98.3} \\
~ & $2 / 4$ & 97.4 & \textbf{98.4} & 97.9 \\
~ & $2 / 9$ & 93.0 & 88.8 & 90.9 \\
~ & $2 / 15$ & 91.2 & 92.8 & 92.0 \\
\cline{2-5}
~ & $3 / 1$ & 97.2 & 94.4 & 95.8 \\
~ & $3 / 3$ & \textbf{99.2} & \textbf{98.8} & \textbf{99.0} \\
~ & $3 / 4$ & 97.2 & 96.0 & 96.6 \\
~ & $3 / 9$ & 92.6 & 88.6 & 90.6 \\
~ & $3 / 15$ & 93.4 & 85.4 & 89.4 \\
\shline 
\label{tab:3}
\end{tabular}
\end{table}

The RAGT architecture has achieved superior results, which we attribute to four factors: 1) The RAGT architecture outputs a $13\times13$ grid, in contrast to the $26\times26$ grid of RARA and RAST. This leads to a more balanced distribution of positive and negative samples in RAGT, thereby enhancing training stability. 2) Within the output grid of RAGT, each grid cell inherently possesses a larger receptive field due to increased downsampling steps. 3) The $13\times13$ grid of RAGT is already sufficiently fine-grained for accurate localization, and using a $26\times26$ grid is unnecessary. 4) RAGT utilizes a global self-attention mechanism that spans across all tokens in the feature maps, enabling the model's receptive field to achieve comprehensive coverage of the entire image. It is conjectured that the overall performance of the RAST architecture surpasses that of the RARA architecture, possibly due to the global receptive field provided by the self-attention mechanism.

\begin{figure}[htbp]
    \centering
    \includegraphics[width=0.45\textwidth]{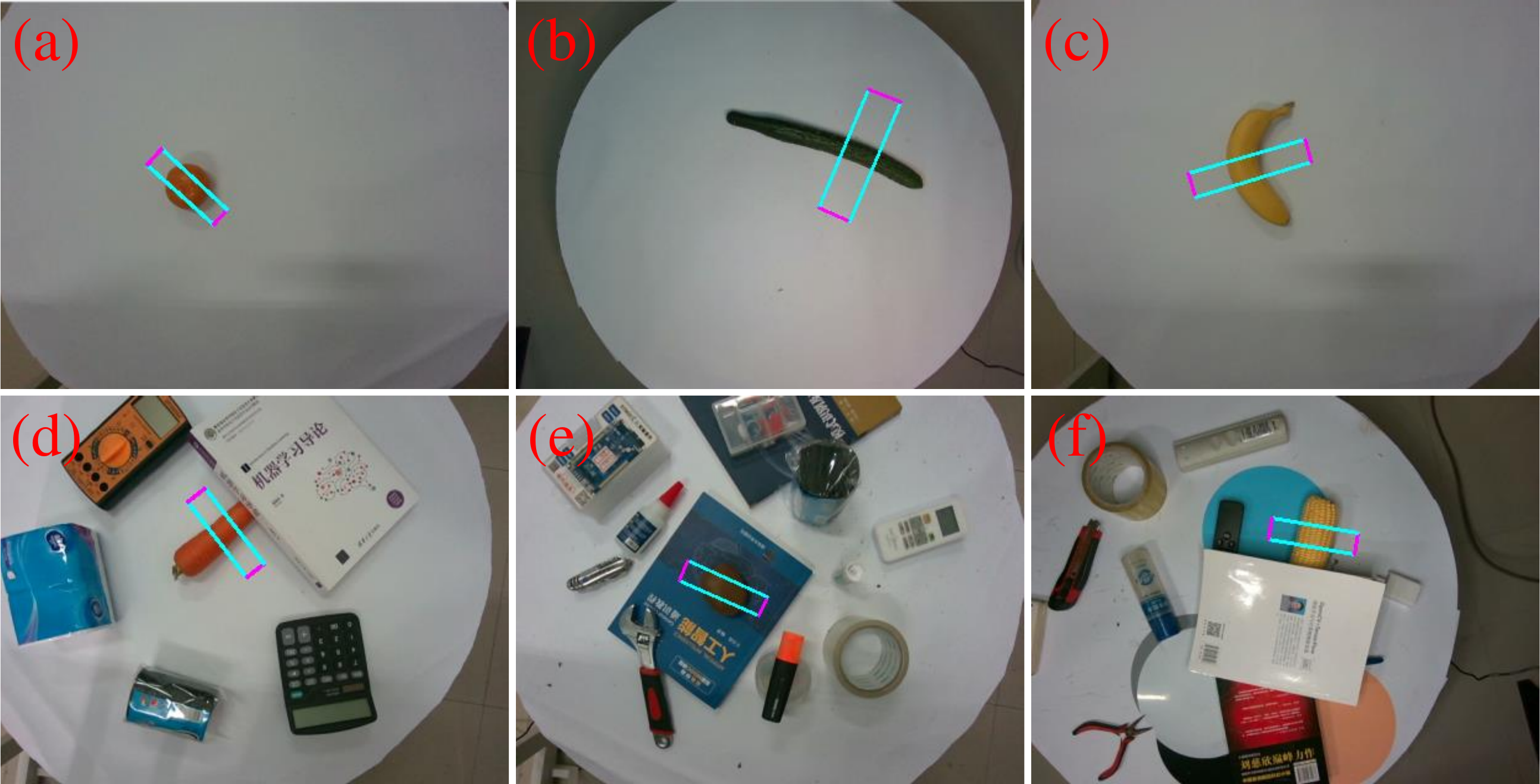}
    \caption{Correct detection results of the RAGT-3/3 model.}
    \label{fig:7}
\end{figure}

\begin{figure}[htbp]
    \centering
    \includegraphics[width=0.45\textwidth]{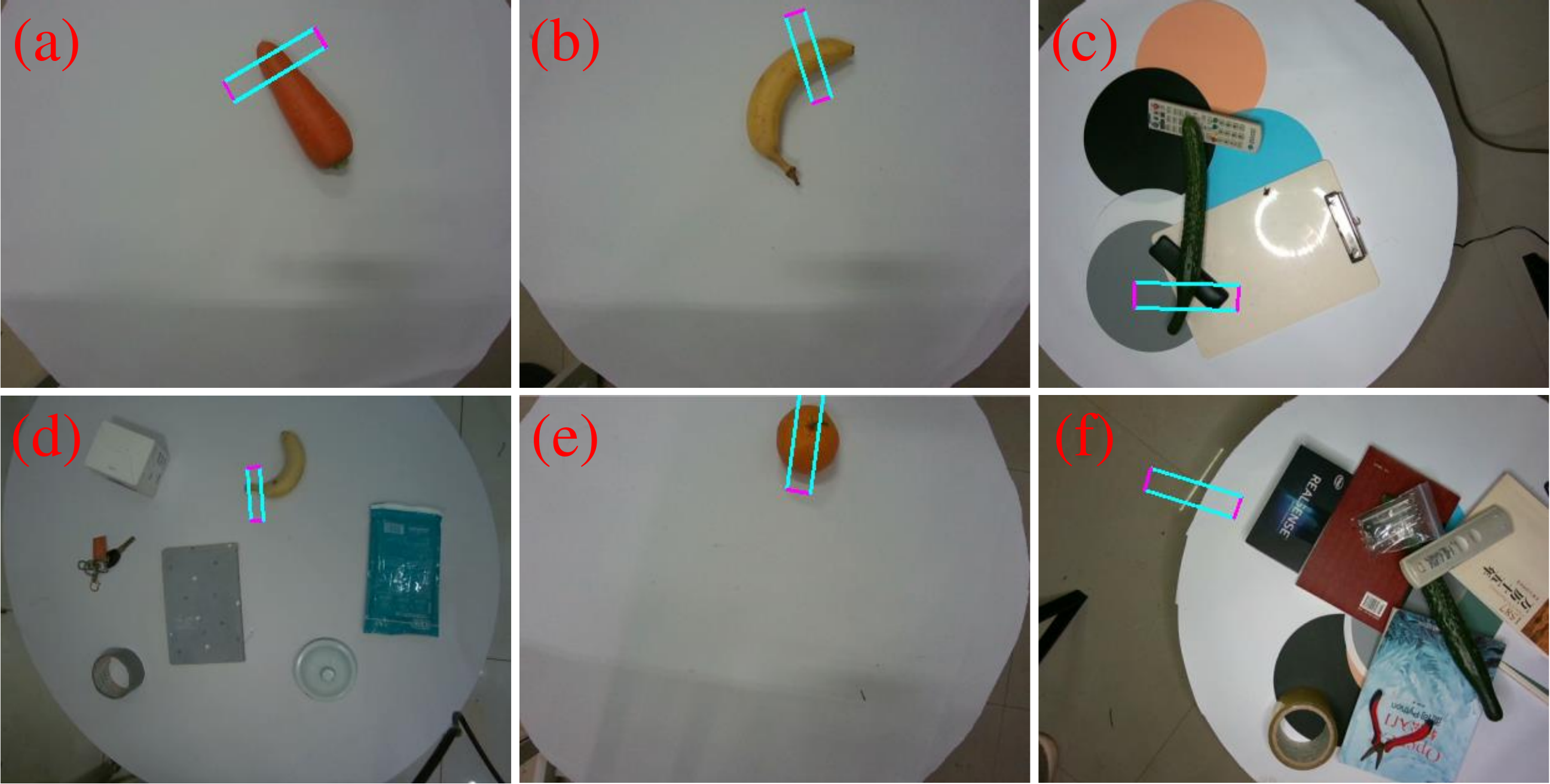}
    \caption{Incorrect detection results of the RAGT-3/3 model.}
    \label{fig:8}
\end{figure}

\par
Taking the RAGT-3/3 model as an example, the correct detection results of this model are shown in Fig.~\ref{fig:7}, while the incorrect detection results are shown in Fig.~\ref{fig:8}. Among the samples identified as incorrect in Fig.~\ref{fig:8}, most of the detected positions could be grasped, such as (a), (b),\textbf{ }(c), (d). The common characteristic of these positions is that the target objects appear in a bar shape, and the predicted grasp positions are located at the ends of the objects, resulting in being judged as ungraspable. For bar-shaped objects, it is desirable for the predicted grasp positions to be as close to the center of the objects as possible. There are two approaches to address this issue: 1) modifying the labels of the model to annotate only the positions near the center of the objects; 2) assigning higher confidence (e.g., 1) to positions near the center of gravity and lower confidence (e.g., 0.6) to positions slightly farther from the center of gravity, while applying an appropriate decay to the regression loss (e.g., multiplying it by the annotated confidence). Although solution 1 is direct, modifying the labels requires significant resources, which is unnecessary. As we have simultaneously annotated the names of each target during image annotation, solution 2 is actually effective and feasible. Sample (e) was deemed incorrect possibly due to the bounding box extending beyond the image, resulting in a low IoU score. The error in sample (f) could be attributed to the mistaken recognition of the lamp's reflection as the target object by the model.

\subsection{Model Sensitivity Analysis}

\begin{table*}[htbp]
\centering
\caption{The results of changing the angular threshold: RARA-3 and RAST-9.}
\begin{tabular}{lccccccc}
\shline
\multirow{2}{*}{Angle threshold} & \multicolumn{3}{c}{RARA-3} & \multicolumn{3}{c}{RAST-9}\\
\cline{2-7} & SSS (\%) & NSS (\%) & Mean (\%) & SSS (\%) & NSS (\%) & Mean (\%) \\
\midrule
$30^{\circ}$ & 98.2 & 94.8 & 96.5 & 97.6 & 97.2 & 97.4 \\
$25^{\circ}$ & 98.0 & 94.8 & 96.4 & 97.2 & 97.2 & 97.2 \\
$20^{\circ}$ & 97.8 & 94.6 & 96.2 & 97.2 & 96.6 & 96.9 \\
$15^{\circ}$ & 97.0 & 92.0 & 94.5 & 96.4 & 94.0 & 95.2 \\
$10^{\circ}$ & 85.4 & 77.2 & 81.3 & 86.8 & 80.6 & 83.7 \\
\shline 
\label{tab:4}
\end{tabular}
\end{table*}

\begin{table*}[htbp]
\centering
\caption{The results of changing the angular threshold: RAGT-2/3 and RAGT-3/3.}
\begin{tabular}{lccccccc}
\shline
\multirow{2}{*}{Angle threshold} & \multicolumn{3}{c}{RAGT-2/3} & \multicolumn{3}{c}{RAGT-3/3}\\
\cline{2-7} & SSS (\%) & NSS (\%) & Mean (\%) & SSS (\%) & NSS (\%) & Mean (\%) \\
\midrule
$30^{\circ}$ & 98.6 & 98.0 & 98.3 & 99.2 & 98.8 & 99.0 \\
$25^{\circ}$ & 98.6 & 98.0 & 98.3 & 99.0 & 98.6 & 98.8 \\
$20^{\circ}$ & 98.6 & 98.0 & 98.3 & 99.0 & 98.6 & 98.8 \\
$15^{\circ}$ & 98.0 & 95.2 & 96.6 & 98.6 & 95.4 & 97.0 \\
$10^{\circ}$ & 85.6 & 80.0 & 82.8 & 84.0 & 78.8 & 81.4 \\
\shline 
\label{tab:5}
\end{tabular}
\end{table*}

\begin{table*}[htbp]
\centering
\caption{The results of changing the IoU threshold: RARA-3 and RAST-9.}
\begin{tabular}{lccccccc}
\shline
\multirow{2}{*}{Angle threshold} & \multicolumn{3}{c}{RAGT-2/3} & \multicolumn{3}{c}{RAGT-3/3}\\
\cline{2-7} & SSS (\%) & NSS (\%) & Mean (\%) & SSS (\%) & NSS (\%) & Mean (\%) \\
\midrule
0.25 & 98.2 & 94.8 & 96.5 & 97.6 & 97.2 & 97.4 \\
0.30 & 96.0 & 94.2 & 95.1 & 95.6 & 97.2 & 96.4 \\
0.35 & 94.6 & 93.0 & 93.8 & 94.8 & 96.4 & 95.6 \\
0.40 & 93.4 & 92.0 & 92.7 & 93.2 & 92.8 & 93.0 \\
0.45 & 91.0 & 87.8 & 89.4 & 92.0 & 90.2 & 91.1 \\
\shline 
\label{tab:6}
\end{tabular}
\end{table*}

\begin{table*}[htbp]
\centering
\caption{The results of changing the IoU threshold: RAGT-2/3 and RAGT-3/3.}
\begin{tabular}{lccccccc}
\shline
\multirow{2}{*}{Angle threshold} & \multicolumn{3}{c}{RAGT-2/3} & \multicolumn{3}{c}{RAGT-3/3}\\
\cline{2-7} & SSS (\%) & NSS (\%) & Mean (\%) & SSS (\%) & NSS (\%) & Mean (\%) \\
\midrule
0.25 & 98.6 & 98.0 & 98.3 & 99.2 & 98.8 & 99.0 \\
0.30 & 97.6 & 97.8 & 97.7 & 98.8 & 98.0 & 98.4 \\
0.35 & 97.2 & 97.8 & 97.5 & 98.2 & 97.8 & 98.0 \\
0.40 & 96.0 & 96.8 & 96.4 & 97.0 & 97.4 & 97.2 \\
0.45 & 94.0 & 94.6 & 94.3 & 95.2 & 95.8 & 95.5 \\
\shline 
\label{tab:7}
\end{tabular}
\end{table*}

In order to further examine the robustness of the model, the evaluation criteria were modified. Specifically, the model was subjected to more stringent evaluation standards. Firstly, it was required that the predicted oriented bounding boxes exhibit smaller angular deviations from the ground truth, while maintaining the unchanged IoU evaluation threshold. The corresponding results are presented in Table~\ref{tab:4} and Table~\ref{tab:5}. Secondly, greater IoU values between the predicted bounding boxes and the ground truth were demanded, while keeping the angular evaluation threshold unchanged. The results are illustrated in Table~\ref{tab:6} and Table~\ref{tab:7}.
\par
Table~\ref{tab:4}, \ref{tab:5}, \ref{tab:6}, and \ref{tab:7} reveal that the utilization of RAM and RA-GraspNet architectures achieves a remarkable accuracy in angle prediction. Even when subjected to stricter angle or IoU thresholds, the model consistently maintains a high level of correctness.

\subsection{Model Attention Analysis}

\begin{figure}[htbp]
    \centering
    \includegraphics[width=0.45\textwidth]{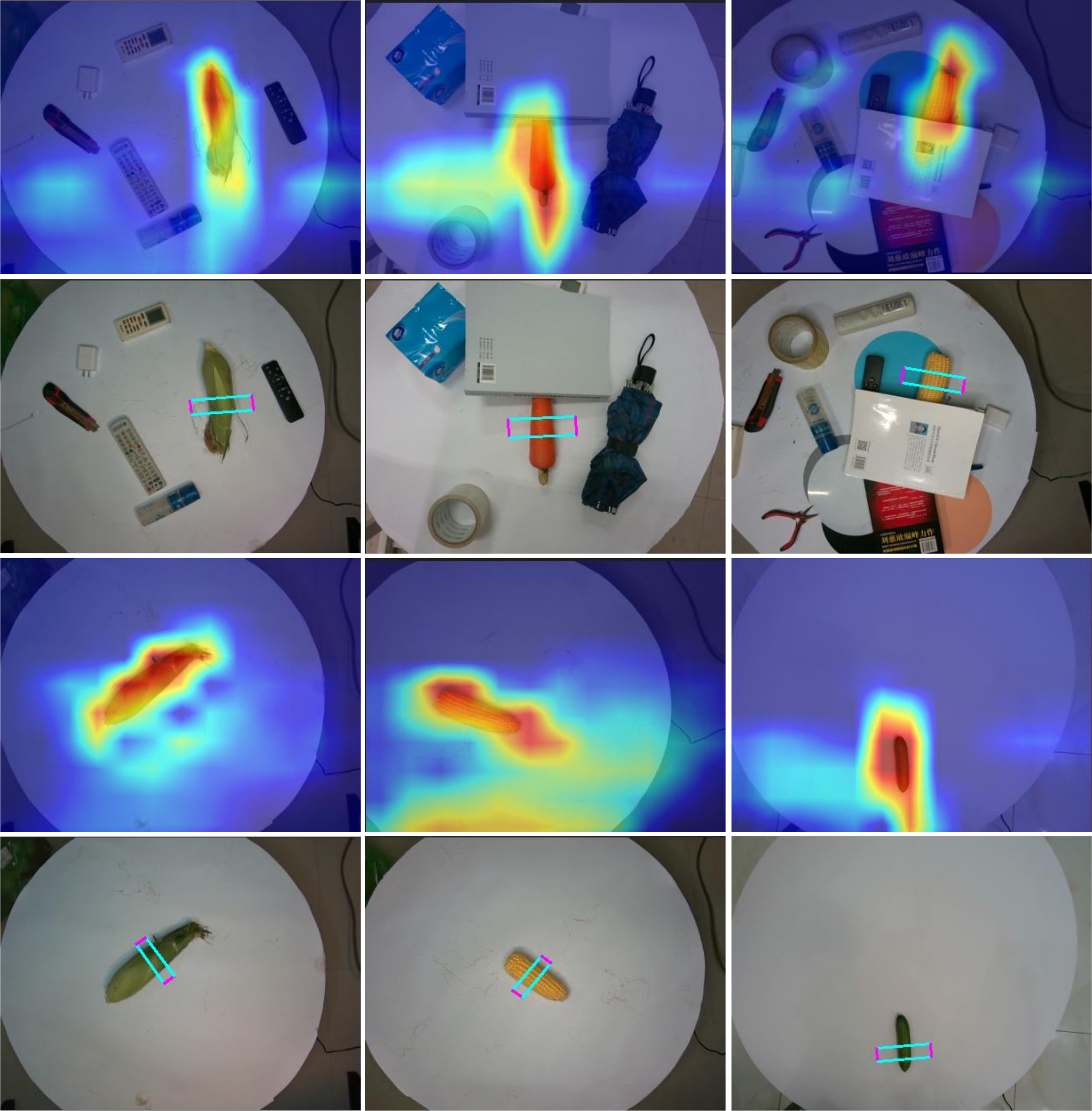}
    \caption{Attention heatmaps of the RAGT-3/3 model.}
    \label{fig:9}
\end{figure}

In order to comprehend how the model makes decisions, Eigen-CAM~\cite{muhammad2020eigen} was employed to visualize the attention distribution of the penultimate layer in the RAGT-3/3 model. The results are illustrated in Fig.~\ref{fig:9}. From the figure, it can be observed that the model has acquired representations concerning the grasping task. Notably, the model demonstrates a remarkable ability to attend to various features such as the position and contours of the target object, as well as the occlusion relationship with obstacles.

\section{Discussion and Limitations}

\subsection{Model Architecture and Semantic Representation}

\begin{figure*}[htbp]
    \centering
    \includegraphics[width=0.95\textwidth]{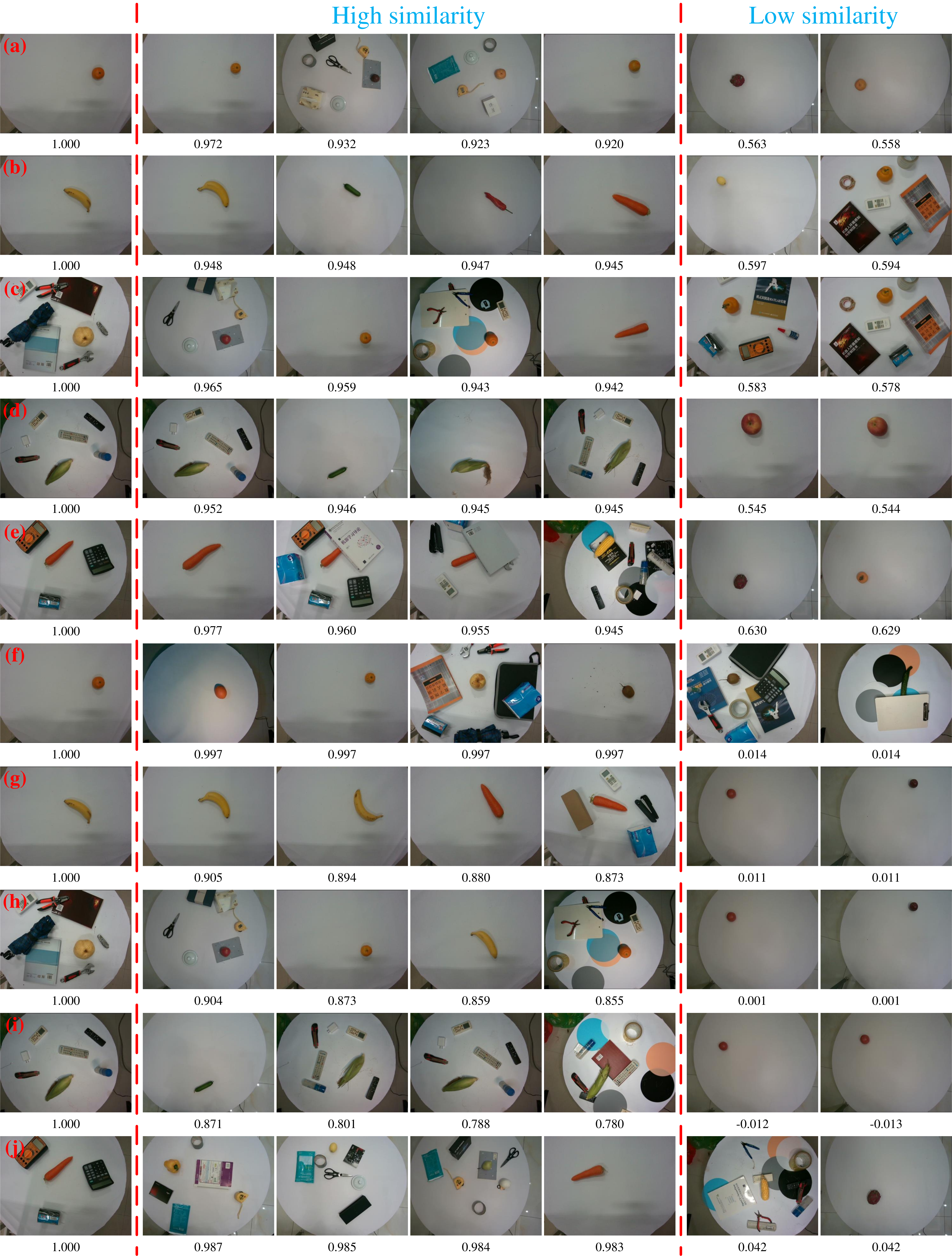}
    \caption{Semantic representation analysis of the RAGT-3/3 model.}
    \label{fig:10}
\end{figure*}

To investigate the representations learned by the model regarding the grasping detection task, we combined the test sets of SSS and NSS. We calculated the output feature maps of the Transformer module in the RAGT-3/3 model for these 1000 samples. Subsequently, we flattened these feature maps into vectors and computed the cosine similarity between these vectors. Fig.~\ref{fig:10} illustrates the images with the highest and lowest similarities to (a), (b), (c), (d), and (e), along with their corresponding similarity scores.
\par
The objects in images with similar feature vectors may not necessarily belong to the same category, and the similarity between images containing objects of the same category is not always high. The model takes into account various features such as the position, pose, and shape of the objects, thus high similarity between images arises from the combined consideration of these features. Objects in images similar to Fig.~\ref{fig:10} (a) may not necessarily be oranges, but they are circular objects located in the upper right part of the image. However, even for circular objects, when there are significant differences in their positions, the similarity to (a) becomes low. Images similar to Fig.~\ref{fig:10} (c) share similar object positions, despite some variations in their shapes. Images similar to Fig.~\ref{fig:10} (e) exhibit objects with similar positions and poses, while samples with low similarity differ significantly from (e) in terms of position and shape. The color of the objects seems to have little impact on grasping detection since the correlation between object color and its suitable grasping position is weak. This indicates that the position, pose, and shape of the target objects are several prominent features that significantly influence the grasping detection task.
\par
Figures (f) to (j) in Fig. 10 replicated the experiments conducted in Figures (a) to (e), with the distinction that the experiments in Figures (f) to (j) utilized the output of the penultimate layer of the RAGT-3/3 model. From Figures (f) to (j), it can be observed that there is no significant variation in the learned representation of the model. However, the cosine similarity of the least similar images is significantly reduced, approaching zero. This indicates that the features of dissimilar images are orthogonal, exhibiting a high degree of irrelevance. Although the residual blocks in the Detector module do not learn additional representations, they further differentiate the semantic space, leading to better clustering of objects with similar semantic concepts and enhancing discrimination of objects with distinct semantic concepts. This suggests that the shallow layers of the model may focus more on representation learning, while the deeper layers increasingly focus on the transformation of the semantic space.

\subsection{Unsupervised Pretraining Methods for Grasp Detection}

In comparison to image classification and object detection, the dataset scale for the grasp detection task is significantly smaller by several orders of magnitude. In scenarios with extremely limited data, utilizing unsupervised learning for pretraining presents an effective strategy to enhance the performance of the model. Based on the preceding analysis, it is evident that the essential characteristics for the grasp detection task encompass the positional, orientational, and geometrical attributes of the target object. Based on these analyses, we propose a feasible unsupervised training approach for grasp detection. Specifically, the method we propose focuses on learning the three aforementioned crucial features.

\begin{itemize}
    \item[1)] \textbf{\textit{In regard to the position information}}, the image can be randomly translated vertically by i pixels and horizontally by j pixels. Subsequently, the original image and the translated image are inputted, and the model is tasked with regressing the values of i and j (with rightward and upward translations defined as positive, and leftward and downward translations defined as negative).
    \item[2)] \textbf{\textit{Regarding the pose information}}, the image can be randomly rotated by an angle, denoted as "$\omega$." The original image and the rotated image are then inputted, and the model is tasked with predicting the rotation angle (with clockwise rotation defined as positive and counterclockwise rotation defined as negative).
    \item[3)] \textbf{\textit{With regard to contour information}}, an unsupervised clustering algorithm, such as KMeans, can be employed to cluster the original image. The original images serve as inputs, while the clustering results act as labels, enabling the model to learn the segmentation task for pretraining purposes.
\end{itemize}

Taking into account the aforementioned three methods, the model's input consists of three types of images: 1) the original image, 2) the image after rotation, and 3) the image after translation. The model's output encompasses four branches: 1) the pixel displacement in the vertical direction, 2) the pixel displacement in the horizontal direction, 3) the rotation angle, and 4) the segmentation result. Using RGB images as an example, during the pretraining stage, the model's input consists of the concatenation of the three images along the channel dimension. Pretraining can be conducted by assigning certain weights to the losses of the four output branches and calculating the overall loss. The issue with this unsupervised pre-training method is that the number of input channels during pre-training differs from those during fine-tuning and inference stages. Two methods were devised to address this issue. 1) Using 9-channel convolutional kernels during pre-training, while in the fine-tuning phase, three parameter channels are randomly selected from the nine channels to form a new convolutional layer for supervised training. 2) Considering the weak correlation between color and grasp detection task, it is possible to convert all RGB images to grayscale as input. This ensures consistency in the number of channels during the pretraining, fine-tuning, and inference stages. By employing the aforementioned unsupervised pretraining methods, large-scale datasets for classification, detection, and segmentation tasks can all be utilized for grasp detection, thus significantly enhancing the robustness of the model.

\subsection{Limitations}

CLIP~\cite{radford2021learning} has gained renown for its remarkable universality, providing a glimmer of hope for "recognizing anything." Recent efforts have increasingly focused on creating a unified model, such as SAM~\cite{kirillov2023segment}. However, the proposed method in this paper still confines the robot system to recognize and grasp only the annotated objects within the training set. Researchers’ ultimate aspiration is for robots to attain a generalized grasping capability, enabling them to grasp anything. By simply providing a prompt, the robot would accurately identify the corresponding object and locate an appropriate position to grasp it. Achieving this goal may entail referencing current multimodal works and establishing a multimodal grasping dataset for fine-tuning purposes.

\section{Conclusion}

This paper presents a noise-scene multi-object grasping detection dataset, named NBMOD, comprising 31,500 RGB-D images. The dataset encompasses 13,500 samples with simple backgrounds, 13,000 samples with noisy backgrounds, and 5,000 multi-object samples. This dataset provides essential material for research on grasping detection in complex environments for robots. A novel oriented bounding box localization mechanism, called RAM, is proposed in this paper. The method serves as a plug-and-play module that can be extended for oriented bounding box localization tasks in remote sensing detection and similar domains. The RA-GraspNet series architectures are introduced: RARA, RAST, and RAGT. Notably, the RAGT-3/3 model achieves a remarkable accuracy of 99\% on the NBMOD dataset. The RA-GraspNet series architectures can be effectively extended to other oriented bounding box localization tasks as well, utilizing multi-branch and multi-scale detection heads when required. Through analysis, we observed that the position, pose, and contour of the target object are three significant features that affect grasp detection tasks. Focusing on these three features, we envisaged a feasible unsupervised training method to pretrain the model to improve its robustness.
\section*{Acknowledgement}

This work was supported by the Jiangsu Agricultural Science and Technology Innovation Fund (JASTIF) (Grant no.  CX (21) 3146).



\begin{thebibliography}{10}
\providecommand{\url}[1]{#1}
\csname url@samestyle\endcsname
\providecommand{\newblock}{\relax}
\providecommand{\bibinfo}[2]{#2}
\providecommand{\BIBentrySTDinterwordspacing}{\spaceskip=0pt\relax}
\providecommand{\BIBentryALTinterwordstretchfactor}{4}
\providecommand{\BIBentryALTinterwordspacing}{\spaceskip=\fontdimen2\font plus
\BIBentryALTinterwordstretchfactor\fontdimen3\font minus
  \fontdimen4\font\relax}
\providecommand{\BIBforeignlanguage}[2]{{%
\expandafter\ifx\csname l@#1\endcsname\relax
\typeout{** WARNING: IEEEtran.bst: No hyphenation pattern has been}%
\typeout{** loaded for the language `#1'. Using the pattern for}%
\typeout{** the default language instead.}%
\else
\language=\csname l@#1\endcsname
\fi
#2}}
\providecommand{\BIBdecl}{\relax}
\BIBdecl

\bibitem{krizhevsky2017imagenet}
A.~Krizhevsky, I.~Sutskever, and G.~E. Hinton, ``Imagenet classification with
  deep convolutional neural networks,'' \emph{Communications of the ACM},
  vol.~60, no.~6, pp. 84--90, 2017.

\bibitem{zeiler2014visualizing}
M.~D. Zeiler and R.~Fergus, ``Visualizing and understanding convolutional
  networks,'' in \emph{Computer Vision--ECCV 2014: 13th European Conference,
  Zurich, Switzerland, September 6-12, 2014, Proceedings, Part I 13}.\hskip 1em
  plus 0.5em minus 0.4em\relax Springer, 2014, pp. 818--833.

\bibitem{simonyan2014very}
K.~Simonyan and A.~Zisserman, ``Very deep convolutional networks for
  large-scale image recognition,'' \emph{arXiv preprint arXiv:1409.1556}, 2014.

\bibitem{girshick2014rich}
R.~Girshick, J.~Donahue, T.~Darrell, and J.~Malik, ``Rich feature hierarchies
  for accurate object detection and semantic segmentation,'' in
  \emph{Proceedings of the IEEE conference on computer vision and pattern
  recognition}, 2014, pp. 580--587.

\bibitem{girshick2015fast}
R.~Girshick, ``Fast r-cnn,'' in \emph{Proceedings of the IEEE international
  conference on computer vision}, 2015, pp. 1440--1448.

\bibitem{ren2015faster}
S.~Ren, K.~He, R.~Girshick, and J.~Sun, ``Faster r-cnn: Towards real-time
  object detection with region proposal networks,'' \emph{Advances in neural
  information processing systems}, vol.~28, 2015.

\bibitem{redmon2016you}
J.~Redmon, S.~Divvala, R.~Girshick, and A.~Farhadi, ``You only look once:
  Unified, real-time object detection,'' in \emph{Proceedings of the IEEE
  conference on computer vision and pattern recognition}, 2016, pp. 779--788.

\bibitem{redmon2017yolo9000}
J.~Redmon and A.~Farhadi, ``Yolo9000: better, faster, stronger,'' in
  \emph{Proceedings of the IEEE conference on computer vision and pattern
  recognition}, 2017, pp. 7263--7271.

\bibitem{redmon2018yolov3}
------, ``Yolov3: An incremental improvement,'' \emph{arXiv preprint
  arXiv:1804.02767}, 2018.

\bibitem{tian2019fcos}
Z.~Tian, C.~Shen, H.~Chen, and T.~He, ``Fcos: Fully convolutional one-stage
  object detection,'' in \emph{Proceedings of the IEEE/CVF international
  conference on computer vision}, 2019, pp. 9627--9636.

\bibitem{liu2016ssd}
W.~Liu, D.~Anguelov, D.~Erhan, C.~Szegedy, S.~Reed, C.-Y. Fu, and A.~C. Berg,
  ``Ssd: Single shot multibox detector,'' in \emph{Computer Vision--ECCV 2016:
  14th European Conference, Amsterdam, The Netherlands, October 11--14, 2016,
  Proceedings, Part I 14}.\hskip 1em plus 0.5em minus 0.4em\relax Springer,
  2016, pp. 21--37.

\bibitem{lin2017feature}
T.-Y. Lin, P.~Doll{\'a}r, R.~Girshick, K.~He, B.~Hariharan, and S.~Belongie,
  ``Feature pyramid networks for object detection,'' in \emph{Proceedings of
  the IEEE conference on computer vision and pattern recognition}, 2017, pp.
  2117--2125.

\bibitem{lin2017focal}
T.-Y. Lin, P.~Goyal, R.~Girshick, K.~He, and P.~Doll{\'a}r, ``Focal loss for
  dense object detection,'' in \emph{Proceedings of the IEEE international
  conference on computer vision}, 2017, pp. 2980--2988.

\bibitem{he2015spatial}
K.~He, X.~Zhang, S.~Ren, and J.~Sun, ``Spatial pyramid pooling in deep
  convolutional networks for visual recognition,'' \emph{IEEE transactions on
  pattern analysis and machine intelligence}, vol.~37, no.~9, pp. 1904--1916,
  2015.

\bibitem{vaswani2017attention}
A.~Vaswani, N.~Shazeer, N.~Parmar, J.~Uszkoreit, L.~Jones, A.~N. Gomez,
  {\L}.~Kaiser, and I.~Polosukhin, ``Attention is all you need,''
  \emph{Advances in neural information processing systems}, vol.~30, 2017.

\bibitem{dosovitskiy2020image}
A.~Dosovitskiy, L.~Beyer, A.~Kolesnikov, D.~Weissenborn, X.~Zhai,
  T.~Unterthiner, M.~Dehghani, M.~Minderer, G.~Heigold, S.~Gelly \emph{et~al.},
  ``An image is worth 16x16 words: Transformers for image recognition at
  scale,'' \emph{arXiv preprint arXiv:2010.11929}, 2020.

\bibitem{liu2021swin}
Z.~Liu, Y.~Lin, Y.~Cao, H.~Hu, Y.~Wei, Z.~Zhang, S.~Lin, and B.~Guo, ``Swin
  transformer: Hierarchical vision transformer using shifted windows,'' in
  \emph{Proceedings of the IEEE/CVF international conference on computer
  vision}, 2021, pp. 10\,012--10\,022.

\bibitem{mehta2021mobilevit}
S.~Mehta and M.~Rastegari, ``Mobilevit: light-weight, general-purpose, and
  mobile-friendly vision transformer,'' \emph{arXiv preprint arXiv:2110.02178},
  2021.

\bibitem{lenz2015deep}
I.~Lenz, H.~Lee, and A.~Saxena, ``Deep learning for detecting robotic grasps,''
  \emph{The International Journal of Robotics Research}, vol.~34, no. 4-5, pp.
  705--724, 2015.

\bibitem{song2020novel}
Y.~Song, L.~Gao, X.~Li, and W.~Shen, ``A novel robotic grasp detection method
  based on region proposal networks,'' \emph{Robotics and Computer-Integrated
  Manufacturing}, vol.~65, p. 101963, 2020.

\bibitem{redmon2015real}
J.~Redmon and A.~Angelova, ``Real-time grasp detection using convolutional
  neural networks,'' in \emph{2015 IEEE international conference on robotics
  and automation (ICRA)}.\hskip 1em plus 0.5em minus 0.4em\relax IEEE, 2015,
  pp. 1316--1322.

\bibitem{he2016deep}
K.~He, X.~Zhang, S.~Ren, and J.~Sun, ``Deep residual learning for image
  recognition,'' in \emph{Proceedings of the IEEE conference on computer vision
  and pattern recognition}, 2016, pp. 770--778.

\bibitem{kumra2017robotic}
S.~Kumra and C.~Kanan, ``Robotic grasp detection using deep convolutional
  neural networks,'' in \emph{2017 IEEE/RSJ International Conference on
  Intelligent Robots and Systems (IROS)}.\hskip 1em plus 0.5em minus
  0.4em\relax IEEE, 2017, pp. 769--776.

\bibitem{dai2017deformable}
J.~Dai, H.~Qi, Y.~Xiong, Y.~Li, G.~Zhang, H.~Hu, and Y.~Wei, ``Deformable
  convolutional networks,'' in \emph{Proceedings of the IEEE international
  conference on computer vision}, 2017, pp. 764--773.

\bibitem{zhou2018fully}
X.~Zhou, X.~Lan, H.~Zhang, Z.~Tian, Y.~Zhang, and N.~Zheng, ``Fully
  convolutional grasp detection network with oriented anchor box,'' in
  \emph{2018 IEEE/RSJ International Conference on Intelligent Robots and
  Systems (IROS)}.\hskip 1em plus 0.5em minus 0.4em\relax IEEE, 2018, pp.
  7223--7230.

\bibitem{ding2022scaling}
X.~Ding, X.~Zhang, J.~Han, and G.~Ding, ``Scaling up your kernels to 31x31:
  Revisiting large kernel design in cnns,'' in \emph{Proceedings of the
  IEEE/CVF Conference on Computer Vision and Pattern Recognition}, 2022, pp.
  11\,963--11\,975.

\bibitem{dong2022robotic}
M.~Dong, Y.~Bai, S.~Wei, and X.~Yu, ``Robotic grasp detection based on
  transformer,'' in \emph{Intelligent Robotics and Applications: 15th
  International Conference, ICIRA 2022, Harbin, China, August 1--3, 2022,
  Proceedings, Part IV}.\hskip 1em plus 0.5em minus 0.4em\relax Springer, 2022,
  pp. 437--448.

\bibitem{ribeiro2019fast}
E.~G. Ribeiro and V.~Grassi, ``Fast convolutional neural network for real-time
  robotic grasp detection,'' in \emph{2019 19th International Conference on
  Advanced Robotics (ICAR)}.\hskip 1em plus 0.5em minus 0.4em\relax IEEE, 2019,
  pp. 49--54.

\bibitem{chu2018deep}
F.-J. Chu and P.~A. Vela, ``Deep grasp: Detection and localization of grasps
  with deep neural networks,'' \emph{arXiv preprint arXiv:1802.00520}, vol.~1,
  2018.

\bibitem{wang2022transformer}
S.~Wang, Z.~Zhou, and Z.~Kan, ``When transformer meets robotic grasping:
  Exploits context for efficient grasp detection,'' \emph{IEEE Robotics and
  Automation Letters}, vol.~7, no.~3, pp. 8170--8177, 2022.

\bibitem{park2018real}
D.~Park, Y.~Seo, and S.~Y. Chun, ``Real-time, highly accurate robotic grasp
  detection using fully convolutional neural networks with high-resolution
  images,'' \emph{arXiv preprint arXiv:1809.05828}, 2018.

\bibitem{depierre2002optimizing}
A.~Depierre, E.~Dellandr{\'e}a, and L.~Chen, ``Optimizing correlated
  graspability score and grasp regression for better grasp prediction. arxiv
  2020,'' \emph{arXiv preprint arXiv:2002.00872}.

\bibitem{caldera2018robotic}
S.~Caldera, A.~Rassau, and D.~Chai, ``Robotic grasp pose detection using deep
  learning,'' in \emph{2018 15th International Conference on Control,
  Automation, Robotics and Vision (ICARCV)}.\hskip 1em plus 0.5em minus
  0.4em\relax IEEE, 2018, pp. 1966--1972.

\bibitem{liu2022pegg}
Z.~Liu, H.~Wang, L.~Zhou, H.~Yin, and M.~H. Ang~Jr, ``Pegg-net: Background
  agnostic pixel-wise efficient grasp generation under closed-loop
  conditions,'' \emph{arXiv preprint arXiv:2203.16301}, 2022.

\bibitem{wang2020sgdn}
D.~Wang, ``Sgdn: Segmentation-based grasp detection network for unsymmetrical
  three-finger gripper,'' \emph{arXiv preprint arXiv:2005.08222}, 2020.

\bibitem{guo2016object}
D.~Guo, T.~Kong, F.~Sun, and H.~Liu, ``Object discovery and grasp detection
  with a shared convolutional neural network,'' in \emph{2016 IEEE
  International Conference on Robotics and Automation (ICRA)}.\hskip 1em plus
  0.5em minus 0.4em\relax IEEE, 2016, pp. 2038--2043.

\bibitem{xu2019graspcnn}
Y.~Xu, L.~Wang, A.~Yang, and L.~Chen, ``Graspcnn: Real-time grasp detection
  using a new oriented diameter circle representation,'' \emph{IEEE Access},
  vol.~7, pp. 159\,322--159\,331, 2019.

\bibitem{mahler2019learning}
J.~Mahler, M.~Matl, V.~Satish, M.~Danielczuk, B.~DeRose, S.~McKinley, and
  K.~Goldberg, ``Learning ambidextrous robot grasping policies,'' \emph{Science
  Robotics}, vol.~4, no.~26, p. eaau4984, 2019.

\bibitem{mahler2017dex}
J.~Mahler, J.~Liang, S.~Niyaz, M.~Laskey, R.~Doan, X.~Liu, J.~A. Ojea, and
  K.~Goldberg, ``Dex-net 2.0: Deep learning to plan robust grasps with
  synthetic point clouds and analytic grasp metrics,'' \emph{arXiv preprint
  arXiv:1703.09312}, 2017.

\bibitem{mahler2018dex}
J.~Mahler, M.~Matl, X.~Liu, A.~Li, D.~Gealy, and K.~Goldberg, ``Dex-net 3.0:
  Computing robust vacuum suction grasp targets in point clouds using a new
  analytic model and deep learning,'' in \emph{2018 IEEE International
  Conference on robotics and automation (ICRA)}.\hskip 1em plus 0.5em minus
  0.4em\relax IEEE, 2018, pp. 5620--5627.

\bibitem{guo2017hybrid}
D.~Guo, F.~Sun, H.~Liu, T.~Kong, B.~Fang, and N.~Xi, ``A hybrid deep
  architecture for robotic grasp detection,'' in \emph{2017 IEEE International
  Conference on Robotics and Automation (ICRA)}.\hskip 1em plus 0.5em minus
  0.4em\relax IEEE, 2017, pp. 1609--1614.

\bibitem{ma2018shufflenet}
N.~Ma, X.~Zhang, H.-T. Zheng, and J.~Sun, ``Shufflenet v2: Practical guidelines
  for efficient cnn architecture design,'' in \emph{Proceedings of the European
  conference on computer vision (ECCV)}, 2018, pp. 116--131.

\bibitem{muhammad2020eigen}
M.~B. Muhammad and M.~Yeasin, ``Eigen-cam: Class activation map using principal
  components,'' in \emph{2020 International Joint Conference on Neural Networks
  (IJCNN)}.\hskip 1em plus 0.5em minus 0.4em\relax IEEE, 2020, pp. 1--7.

\bibitem{radford2021learning}
A.~Radford, J.~W. Kim, C.~Hallacy, A.~Ramesh, G.~Goh, S.~Agarwal, G.~Sastry,
  A.~Askell, P.~Mishkin, J.~Clark \emph{et~al.}, ``Learning transferable visual
  models from natural language supervision,'' in \emph{International conference
  on machine learning}.\hskip 1em plus 0.5em minus 0.4em\relax PMLR, 2021, pp.
  8748--8763.

\bibitem{kirillov2023segment}
A.~Kirillov, E.~Mintun, N.~Ravi, H.~Mao, C.~Rolland, L.~Gustafson, T.~Xiao,
  S.~Whitehead, A.~C. Berg, W.-Y. Lo \emph{et~al.}, ``Segment anything,''
  \emph{arXiv preprint arXiv:2304.02643}, 2023.

\end{thebibliography}
\end{document}